\documentclass[acmsmall,screen]{acmart}
\AtBeginDocument{%
  \providecommand\BibTeX{{%
    \normalfont B\kern-0.5em{\scshape i\kern-0.25em b}\kern-0.8em\TeX}}}

\setcopyright{acmcopyright}
\copyrightyear{2023}
\acmYear{2022}

\acmJournal{TOIS}
\acmVolume{1}
\acmNumber{1}
\acmArticle{1}
\acmMonth{12}

\usepackage{amsfonts}
\usepackage{threeparttable}
\usepackage{multirow}
\usepackage{multicol}
\usepackage{makecell}
\usepackage{algpseudocode}
\usepackage{booktabs}
\usepackage{graphicx}
\usepackage{subfigure}
\newcommand{\tabincell}[2]{\begin{tabular}{@{}#1@{}}#2\end{tabular}}
\usepackage{algorithm}

\algnewcommand\algorithmicforeach{\textbf{for each}}
\algdef{S}[FOR]{ForEach}[1]{\algorithmicforeach\ #1\ \algorithmicdo}

\begin{document}

\title{Unifying Token and Span Level Supervisions for Few-Shot Sequence Labeling}


\author{Zifeng Cheng}
\authornote{Work is done during internship at Tencent Cloud Xiaowei.}
\email{chengzf@smail.nju.edu.cn}
\affiliation{%
  \thanks{This work is supported by National Natural Science Foundation of China under Grant Nos. 61972192, 62172208, 61906085, 41972111. This work is partially supported by Collaborative Innovation Center of Novel Software Technology and Industrialization.}
  \institution{State Key Laboratory for Novel Software Technology, Nanjing University}
  \streetaddress{163 Xianlin Ave}
  \city{Nanjing}
  \state{Jiangsu}
  \country{China}
  \postcode{210023}
}
\author{Qingyu Zhou}
\email{qyzhgm@gmail.com}
\affiliation{%
  \institution{Tencent Cloud Xiaowei}
  \city{Beijing}
  \country{China}
  \postcode{100000}
}
\author{Zhiwei Jiang}
\authornote{Corresponding Author.}
\email{jzw@nju.edu.cn}
\affiliation{%
  \institution{State Key Laboratory for Novel Software Technology, Nanjing University}
  \streetaddress{163 Xianlin Ave}
  \city{Nanjing}
  \state{Jiangsu}
  \country{China}
  \postcode{210023}
}
\author{Xuemin Zhao}
\email{xmzhao1986@gmail.com}
\affiliation{%
  \institution{Tencent Cloud Xiaowei}
  \city{Chengdu}
  \state{Sichuan}
  \country{China}
  \postcode{610000}
}
\author{Yunbo Cao}
\email{yunbocao@tencent.com}
\affiliation{%
  \institution{Tencent Cloud Xiaowei}
  \city{Beijing}
  \country{China}
  \postcode{100000}
}
\author{Qing Gu}
\email{guq@nju.edu.cn}
\affiliation{%
  \institution{State Key Laboratory for Novel Software Technology, Nanjing University}
  \streetaddress{163 Xianlin Ave}
  \city{Nanjing}
  \state{Jiangsu}
  \country{China}
  \postcode{210023}
}

\renewcommand{\shortauthors}{Cheng, et al.}

\begin{abstract}
Few-shot sequence labeling aims to identify novel classes based on only a few labeled samples.
Existing methods solve the data scarcity problem mainly by designing token-level or span-level labeling models based on metric learning.
However, these methods are only trained at a single granularity (i.e., either token level or span level) and have some weaknesses of the corresponding granularity.
In this paper, we first unify token and span level supervisions and propose a Consistent Dual Adaptive Prototypical (CDAP) network for few-shot sequence labeling.
CDAP contains the token-level and span-level networks, jointly trained at different granularities.
To align the outputs of two networks, we further propose a consistent loss to enable them to learn from each other.
During the inference phase, we propose a consistent greedy inference algorithm that first adjusts the predicted probability and then greedily selects non-overlapping spans with maximum probability.
Extensive experiments show that our model achieves new state-of-the-art results on three benchmark datasets.
All the code and data of this work will be released at \url{https://github.com/zifengcheng/CDAP}.
\end{abstract}



\maketitle

\section{Introduction}
Sequence labeling tasks such as Named Entity Recognition (NER) and Slot Tagging (ST) are fundamental tasks in information extraction, benefiting query and document understanding in information retrieval~\cite{DBLP:journals/tois/GuoCFSZC22,DBLP:journals/tois/AgostiMS20,DBLP:conf/www/LiMMLZMWH21}, request analysis in task-oriented dialog systems~\cite{DBLP:journals/tois/VakulenkoKR21,DBLP:journals/tois/MaLZLL22}, and so forth.
Traditional sequence labeling methods are often supervised methods, which require a large amount of annotated data to train.
However, this limits their applications in some real scenarios, where the acquisition of annotated data is laborious and expensive.

In recent years, few-shot sequence labeling has attracted much attention since it aims to extract novel classes based on only a few labeled samples~\cite{DBLP:conf/sac/FritzlerLK19,DBLP:conf/acl/HouCLZLLL20,DBLP:conf/emnlp/YangK20,DBLP:conf/acl/TongWX0L0L21}.
Table~\ref{example} shows a 2-way 1-shot example, which contains two novel classes (i.e., building-other and building-library) and each class has one sample in the support set.
Few-shot sequence labeling aims to learn a model that borrows the prior experience from old domains and adapts to new domains quickly with only very few samples.
For example, the model should extract ``brooklyn bridge'' and ``Oxford University Museum of Natural History'' as ``building-other'', and ``public library'' and ``Radcliffe Science Library'' as ``building-library'' in the query set based on the support set.

Some prevalent methods of few-shot sequence labeling usually focus on token-level metric learning~\cite{DBLP:conf/sac/FritzlerLK19,DBLP:conf/acl/HouCLZLLL20,DBLP:conf/emnlp/YangK20,DBLP:conf/eacl/OguzV21,ma2022label,DBLP:conf/acl/DasKPZ22,DBLP:journals/tkde/LiCFW22}, in which the model assigns a label to each query token based on a learned distance metric.
For example,~\citet{DBLP:conf/sac/FritzlerLK19} construct prototype for each class to classify query tokens and ~\citet{DBLP:conf/acl/DasKPZ22} optimize token-level Gaussian-distributed embeddings to classify query tokens.
Recently, some methods~\cite{DBLP:conf/naacl/YuHZDPL21,DBLP:conf/emnlp/WangCZG21,DBLP:conf/acl/MaJWZL22,DBLP:conf/naacl/WangXLZCCS22} focus on span-level metric learning to measure span-level similarity for classification and achieve state-of-the-art performance.
For example, \citet{DBLP:conf/emnlp/WangCZG21} propose a two-step model, which first detects span and classifies span based on span-level metric.

\begin{table}[b]
\definecolor{cGreen}{RGB}{41,163,95}
\caption{An example of 2-way 1-shot sequence labeling task. Novel classes are {\color{cGreen}building-other} and {\color{orange}building-library}. It is worth noting that in this example two novel classes have the same sentence in the support set.} \label{example}
\begin{tabular}{p{2.1cm}|p{9cm}}
\toprule
\textbf{Support Set $\mathcal{S}$} &(1) The museum was established as the {\color{cGreen}Municipal Gallery of Hamilton [building-other]} in January 1914, and was opened to the public in June 1914, at a {\color{orange}Hamilton Public Library [building-library]} building on Main Street West.\\
\midrule
\textbf{Query Set $\mathcal{Q}$} &(1) The {\color{cGreen}Brooklyn Bridge [building-other]} ceramic tiles display the bridge's vertical cables but do not depict its diagonal cables. (2) The reading program had been discontinued, library patronage had dropped, and the denver {\color{orange}public library [building-library]} was planning to shutter the facility. (3) The current {\color{orange}Radcliffe Science Library [building-library]} building is located next to the {\color{cGreen}Oxford University Museum of Natural History [building-other]} and consists of three parts.\\
\bottomrule
\end{tabular}
\end{table}

Despite their success, these token-level and span-level metric learning methods are only trained at a single granularity, so they inevitably have some weaknesses specific to their granularity.
For the token-level metric learning methods, their scheme of labeling tokens separately is prone to ignore the integrality of named entities and dialog slots with multiple tokens.
Taking the sentence of the support set in Table~\ref{example} as an example, to extract the complete entity ``Municipal Gallery of Hamilton'', each of its constituent tokens, ``Municipal'', ``Gallery'', ``of'', and ``Hamilton'', must be labeled separately, potentially leading to the loss of information about the entity as a whole.
As a consequence, it is challenging for token-level representation learning to capture the full semantics of multi-token entities and slots~\cite{DBLP:conf/naacl/WangXLZCCS22}.
For the span-level metric learning methods, since they need to enumerate all possible spans for classification, 
the few-shot labels of entities and slots are easy to be diluted by a large amount of non-entity and non-slot labels. 
For example, if we set the maximum span length to 4, the spans containing the word ``Municipal'' include ``Municipal'', ``the Municipal'', ``Municipal Gallery'', ``as the Municipal'', ``Municipal Gallery of'', ``established as the Municipal'', ``Municipal Gallery of Hamilton'', and so forth.
Among these spans, only the span ``Municipal Gallery of Hamilton'' is an entity, while the others are non-entities.
According to our statistics of the real dataset FewNERD, the ratio of entity spans to non-entity spans can even exceed 1:100. 
As a result, when entity spans are rare in the few-shot setting, it becomes challenging for span-level metric learning methods to distinguish them from a large number of similar non-entity spans.

These weaknesses are difficult to deal with when only a single-granularity model is trained, but they are promising to be alleviated when both token-level and span-level labels are utilized in a unified view for model training.
On the one hand, span-level labeling can help token-level labeling to maintain entity integrality and capture the full semantics of multi-token entities.
On the other hand, token-level labeling can in turn help span-level labeling to distinguish easily confused span.

To this end, we propose a Consistent Dual Adaptive Prototypical (CDAP) network for few-shot sequence labeling.
CDAP can jointly train token-level and span-level networks, and make them benefit from each other, to produce better predictions.
Specifically, we employ an Adaptive Prototypical Network (APN) to generate adaptive prototypes specific to each query in the few-shot setting and use the APN for both token-level and span-level labeling.
Besides, considering that token-level and span-level networks may produce different predictions, we further propose a consistent loss to align their predictions based on bidirectional Kullback-Leibler divergence with temperature.
During the inference phase, we propose a consistent greedy inference algorithm to combine the predictions of both networks.
The core idea of the algorithm is to adjust the prediction probability of spans based on the predictions of tokens within the span.

The main contributions of this paper can be summarized as follows:
\begin{itemize}
\item To the best of our knowledge, we first unify token and span level supervisions to train the few-shot sequence labeling model.
\item We propose a consistent dual adaptive prototypical network for few-shot sequence labeling and a consistent greedy inference algorithm to combine the results of two networks.
\item Extensive experiments show that our model achieves new state-of-the-art performance on three benchmark datasets.
\end{itemize}

The rest of this paper is organized as follows.
We first review the related work in Section~\ref{sec:related}. 
Then, we formulate the task and describe our proposed method in detail in Section~\ref{sec:task} and ~\ref{sec:model} respectively.
After that, we conduct experiments and analyze the results in Section~\ref{sec:exp}.
Finally, we conclude our work along with some future work in Section~\ref{sec:conclu}.

\section{Related Work}\label{sec:related}

In this section, we introduce the following three research topics relevant to our work: few-shot learning, sequence labeling, and few-shot sequence labeling.

\subsection{Few-Shot Learning}
Few-shot learning aims to recognize novel classes with only few samples in each class and is proposed in the computer vision community~\cite{DBLP:journals/pami/Fei-FeiFP06}.
Motivated by a simple machine learning principle (i.e., train and test must match), matching Networks~\cite{vinyals2016matching} first models the few-shot learning as an episode N-way K-shot task, where each episode has N classes and each class has K annotated samples (K is very small, e.g., 5).
Afterwards, many methods have been proposed, such as metric learning methods, optimization methods, and generation methods.
Metric learning methods aim to learn a suitable embedding space for distance metrics, including cosine similarity~\cite{vinyals2016matching}, euclidean distance to prototype (i.e., mean class representation)~\cite{snell2017prototypical}, learnable relation modules~\cite{DBLP:conf/cvpr/SungYZXTH18}, nearest neighbor~\cite{DBLP:conf/emnlp/YangK20}, and reachable distance~\cite{zhang2022reachable}.
For example,~\citet{snell2017prototypical} first construct prototypes by averaging the representations of each class in the support set and then optimizes the network by using softmax function over Euclidean distances between all prototypes and query representation.
Some other methods have focused on improving the transferability of features, including using cross-attention module to exploit the semantic relevance between support and query features~\cite{DBLP:conf/nips/HouCMSC19}, adaptive margin loss~\cite{DBLP:conf/cvpr/Li0LFLW20}, and so on.
Recently, some methods~\cite{DBLP:conf/iclr/ChenLKWH19,DBLP:conf/iclr/DhillonCRS20,DBLP:conf/iccv/Chen00D021} propose a transfer-learning framework that pre-trains on the training set and fine-tunes on the support set of test episode.
Model-agnostic meta-learning (MAML)~\cite{DBLP:conf/icml/FinnAL17} is a typical approach of optimization methods and aims to learn a good model initialization and then quickly adapt on test episode.
\citet{DBLP:conf/nips/FinnXL18} further extend MAML, which adapts to new tasks via gradient descent, to incorporate a parameter distribution that is trained via a variational lower bound.
Generation methods aim to generate samples~\cite{DBLP:conf/cvpr/WangGHH18,DBLP:conf/nips/ZhangCGBS18,DBLP:conf/acl/ZhouZTJY22} or features~\cite{DBLP:conf/cvpr/Li20} to augment the training set.
Specifically,~\citet{DBLP:conf/nips/ZhangCGBS18} propose to augment vanilla few-shot classification models with GAN for better generalization.
FlipDA~\cite{DBLP:conf/acl/ZhouZTJY22} jointly uses a generative model and a classifier to generate label-flipped data, which has a more significant performance improvement compared to label-preserved data.
AFHN~\cite{DBLP:conf/cvpr/Li20} generates diverse and discriminative features conditioned on the few labeled samples based on cWGAN.

Recently, few-shot learning has received increasing attention in the IR and NLP communities, including recommendation~\cite{DBLP:conf/www/LiWW020}, document filtering~\cite{DBLP:journals/tois/LiuLZJDC20}, dense retrieval~\cite{DBLP:conf/sigir/YuLXF021}, node classification~\cite{DBLP:conf/sigir/LiuLLGFG22}, and so on.

\subsection{Sequence Labeling}
Sequence labeling task is a fundamental task in the field of NLP such as named entity recognition~\cite{DBLP:conf/naacl/LampleBSKD16}, part-of-speech tagging~\cite{Li15}, slot tagging~\cite{DBLP:conf/emnlp/QinCLWL19}, and so on.
Recently, the neural sequence labeling models become the mainstream method for sequence labeling.

Typical neural sequence labeling models~\cite{Huang15,Chiu16,DBLP:conf/naacl/LampleBSKD16,DBLP:conf/acl/MaH16} generally use CNN or LSTM to extract features and utilize softmax or Conditional Random Field (CRF) for classification.
For example,~\citet{Chiu16} use CNN to extract character-level features, LSTM to extract word-level features, and softmax to classify words.
However, \citet{Cui19} find that BiLSTM-CRF does not always lead to better results compared with BiLSTM-Softmax, and propose a hierarchically-refined label attention network, which explicitly leverages label embeddings and captures potential long-term label dependency by giving each word incrementally refined label distributions with hierarchical attention.
Afterwards, some work~\cite{DBLP:conf/acl/YuBP20,DBLP:conf/acl/FuHL20,DBLP:conf/acl/OuchiSKYKKI20,DBLP:conf/acl/JiangXAN20,DBLP:journals/corr/abs-2210-04182} adopts span-level prediction which first detects spans by enumeration or boundary identification, and then classifies spans.
For example,~\citet{DBLP:conf/acl/YuBP20} use a biaffine classifier to assign scores for all spans in a sentence.
\citet{DBLP:conf/acl/FuHL20} first compare and analyze the advantages and disadvantages of the traditional token-level models and span-level models in detail, and then reveal that span prediction can serve as a system combiner to re-recognize named entities from different systems' outputs.

Besides, four novel paradigms for sequence labeling have recently been proposed, reformulating it as sequence-to-sequence learning, generation, set prediction, and machine reading comprehension tasks, respectively.
First, considering that the sequence-to-sequence learning has a good ability to learn complex global label dependency, some work applies it to solve the sequence labeling tasks such as chunking~\cite{Zhai17}, aspect term extraction~\cite{Ma19}, emotion-cause pair extraction~\cite{DBLP:journals/taslp/ChengJYLG21}, and dialogue act prediction problem~\cite{Colombo20}.
For example, in order to make sequence-to-sequence learning suitable for sequence labeling,~\citet{Ma19} design gated unit networks to incorporate corresponding word representation into the decoder and position-aware attention to pay more attention to the adjacent words of a target word.
Secondly, some work use generation to solve the sequence labeling.
Specifically,~\citet{DBLP:conf/emnlp/AthiwaratkunSKX20} propose to generate augmented output that repeats the original input sequence with additional markers that indicate the token-spans and their associated labels.
\citet{DBLP:conf/acl/YanGDGZQ20} propose a unified generation framework with the pointer mechanism to tackle flat, nested, and discontinuous NER subtasks.
\citet{DBLP:conf/acl/Zhang0TW022} analyze the incorrect biases in the generation process from a causality perspective and design intra- and inter-entity deconfounding data augmentation methods to eliminate confounders.
Thirdly, \citet{DBLP:conf/ijcai/Tan0Z0Z21} first propose a sequence-to-set network for NER, which provides a fixed set of entity queries to replace the explicit candidate spans and uses self-attention to capture the dependencies between entities.
Finally, \citet{DBLP:conf/acl/LiFMHWL20} first propose a machine reading comprehension framework to solve the NER task and use type-specific queries using semantic
prior information for entity categories.
However, type-specific queries can only extract one entity type per inference and rely on external knowledge to construct queries.
\citet{DBLP:conf/acl/Shen22} further set up global and learnable instance queries to extract entities from a sentence in a parallel manner.

\subsection{Few-Shot Sequence Labeling}
Few-shot sequence labeling aims to identify novel classes based on only a few labeled samples and has gained widespread attention due to its application value.
We roughly divide the existing few-shot sequence labeling methods into three groups.

The first group of methods is token-level metric learning method.
\citet{DBLP:conf/sac/FritzlerLK19} first use prototypical networks to solve this task.
Afterwards, L-TapNet~\cite{DBLP:conf/acl/HouCLZLLL20} and StructShot~\cite{DBLP:conf/emnlp/YangK20} both explore CRF to model label dependence in the few-shot setting.
L-TapNet~\cite{DBLP:conf/acl/HouCLZLLL20} extends task-adaptive projection networks (TapNet)~\cite{DBLP:conf/icml/YoonSM19}, which improves label representation with label name.
StructShot~\cite{DBLP:conf/emnlp/YangK20} trains model through traditional cross-entropy loss and uses nearest neighbor for classification.
\citet{DBLP:conf/emnlp/HuangLSJBCPG021} explore noisy supervised pre-training and self-training in the few-shot sequence labeling setting.
Motivated by non-entity class O has rich semantics, MUCO~\cite{DBLP:conf/acl/TongWX0L0L21} introduces a binary classifier to determine whether it is an entity or not.
\citet{ma2022label} propose to leverage the semantics of label names and use two separate BERT to encode label and token for token-level similarity metric.
Instead of optimizing class-specific attributes, CONTAINER~\cite{DBLP:conf/acl/DasKPZ22} models Gaussian embedding for each token and optimizes inter token distribution distance to improve generalizability.

The second group of methods is span-level method.
SpanNER~\cite{DBLP:conf/emnlp/WangCZG21} is a two-step method, which first detects the start and end tokens to extract span and then introduces class descriptions to classify spans.
\citet{DBLP:conf/acl/MaJWZL22} further propose a model-agnostic meta-learning (MAML)~\cite{DBLP:conf/icml/FinnAL17} enhanced two-step method, which uses BIESO scheme to detect span and prototypical networks to classify span.
Retriever~\cite{DBLP:conf/naacl/YuHZDPL21} proposes a span-level retrieval method that learns similar contextualized representations for spans with the same label via a novel batch-softmax objective. 
ESD~\cite{DBLP:conf/naacl/WangXLZCCS22} proposes inter-span and cross-span attention to enhance the span representations, and uses instance span attention to construct prototype.

The third group of methods mitigates data scarcity with other paradigms, including machine reading comprehension~\cite{DBLP:conf/acl/MaYLZ21}, generation~\cite{DBLP:conf/emnlp/AthiwaratkunSKX20,DBLP:conf/acl/ChenLLH022}, pairwise cloze~\cite{DBLP:conf/naacl/HendersonV21}, and prompt tuning~\cite{DBLP:conf/acl/CuiWLYZ21,DBLP:conf/acl/LeeKTAF0MSPR22,DBLP:conf/naacl/MaZGTLZH22,DBLP:conf/acl/HouCLLC22}.
For machine reading comprehension method, \citet{DBLP:conf/acl/MaYLZ21} enumerate all the slots to extract the answer from the sentence.
The MRC-based method can naturally encode the label semantics in the form of questions.
For generation methods, \citet{DBLP:conf/emnlp/AthiwaratkunSKX20} propose to generate augmented natural language output, including the original input sequence, delimiter, and label.
Self-describing Networks~\cite{DBLP:conf/acl/ChenLLH022} is a sequence-to-sequence generation network which can universally describe mentions using concepts, automatically map novel entity types to concepts, and adaptively recognize entities on-demand.
For prompt tuning methods, Template~\cite{DBLP:conf/acl/CuiWLYZ21} treats NER as a language model ranking problem in a sequence-to-sequence framework, where the output sequence is filled by templates.
\citet{DBLP:conf/acl/HouCLLC22} propose an inverse prompt, predicting slot values given slot types, and further introduce the iterative prediction strategy to consider the relations between different slot types.
\citet{DBLP:conf/naacl/MaZGTLZH22} reformulate the NER task as an entity-oriented LM task, inducing the LM to predict label words at entity positions during fine-tuning, and propose a template-free prompt tuning method EntLM.
\citet{DBLP:conf/acl/LeeKTAF0MSPR22} perform a systematic study of entity-oriented prompt (i.e., only entity tokens) and instance-oriented prompt (i.e., retrieved sentences).

\section{Task Formulation} \label{sec:task}
We define a sentence as $\boldsymbol{x} = \{x_1, x_2, ..., x_n\}$ and its label as $\boldsymbol{y} = \{(s_i,y_i)\}_{i=1}^M$, where $s_i$ is a span in sentence $\boldsymbol{x}$ (e.g., ``nimbus'' and ``nimbus powerplant'' in Figure~\ref{fig:framework}), $y_i$ is the corresponding label (e.g., entity class ``building-other'' and non-entity class O), and $M$ is the number of spans in the sentence.
Few-shot sequence labeling aims to train a model in the source domain that can accurately extract and classify all novel classes with a few labeled samples in the target domain.
Episode training~\cite{vinyals2016matching} is a common and effective way to train the model by imitating test situation for few-shot learning.
Each training/test episode contains a $N$-way $K$-shot support set $\mathcal{S} = \{(\boldsymbol{x}_i, \boldsymbol{y}_i)\}_{i=1}^{m_s}$ and a query set $\mathcal{Q} = \{(\boldsymbol{x}_i, \boldsymbol{y}_i)\}_{i=1}^{m_q}$, where the support set $\mathcal{S}$ includes $N$ classes (N-way) and each class has K annotated examples (K-shot).
In the training phase, few-shot sequence labeling model is trained by predicting labels of the query set $\mathcal{Q}_{train}$ according to a few labeled support set $\mathcal{S}_{train}$.
In the testing phase, the model directly predicts labels of the query set $\mathcal{Q}_{test}$ on unseen test dataset according to a few labeled support set $\mathcal{S}_{test}$.
To facilitate the illustration, we list the main notations used throughout this article in Table~\ref{notation}.

\begin{table}[h]
\centering
\caption{Main Notations Used in this Article.}\label{notation}
\begin{tabular}{c|l}
\toprule
\textbf{Notation} & \textbf{Description}\\
\midrule
$\mathbf{H} \in \mathbb{R}^{n \times d}$ & Token-level representation matrix for a sentence $\boldsymbol{x}$ \\
$\mathbf{H^s_j} \in \mathbb{R}^{C_{j} \times d}$  & Representation matrix of all tokens of class $j$ in the support set\\
$\mathbf{h_i^q} \in \mathbb{R}^{d}$ &  Token-level representation for query token $x_i$\\
$\mathbf{c_i^j} \in \mathbb{R}^{d}$ &  Token-level adaptive prototype of class $j$ for query token $x_i$\\
$n$ &  The length of sentence $\boldsymbol{x}$\\
$d$ &  The dimension of feature representation\\
$C_j$ &  The number of tokens of class $j$ in the support set\\
$\mathbf{S} \in \mathbb{R}^{S_s \times d}$ & Representation matrix of all spans in the support set\\
$\mathbf{Q} \in \mathbb{R}^{S_q \times d}$ & Representation matrix of all spans in the query set\\
$\bar{\mathbf{S}} \in \mathbb{R}^{S_s \times d}$ & Representation matrix of all spans after cross-attention module in the support set\\
$\bar{\mathbf{Q}} \in \mathbb{R}^{S_q \times d}$ & Representation matrix of all spans after cross-attention module in the query set\\
$\hat{\textbf{o}}_k \in \mathbb{R}^{d}$ & Span-level adaptive prototype of sub-class $O_k$ for query span $q_i$\\
$\textbf{O}_k \in \mathbb{R}^{o_{k}\times d}$ & Representation matrix of all spans in the support set with sub-class $O_k$\\
$\bar{\textbf{q}}_i \in \mathbb{R}^{d}$ &  Span-level representation for query span $q_i$\\
$\mathbf{z_i^j} \in \mathbb{R}^{d}$ &  Span-level adaptive prototype of class $j$ for query span $q_i$\\
$S_s$ & The number of spans in the support set\\
$S_q$ & The number of spans in the query set\\
$o_k$ & The number of spans with sub-class $O_k$ in the support set\\
\bottomrule
\end{tabular}
\end{table}

\section{Model} \label{sec:model}
\begin{figure*}[t]
\centering
\includegraphics[width=1\columnwidth]{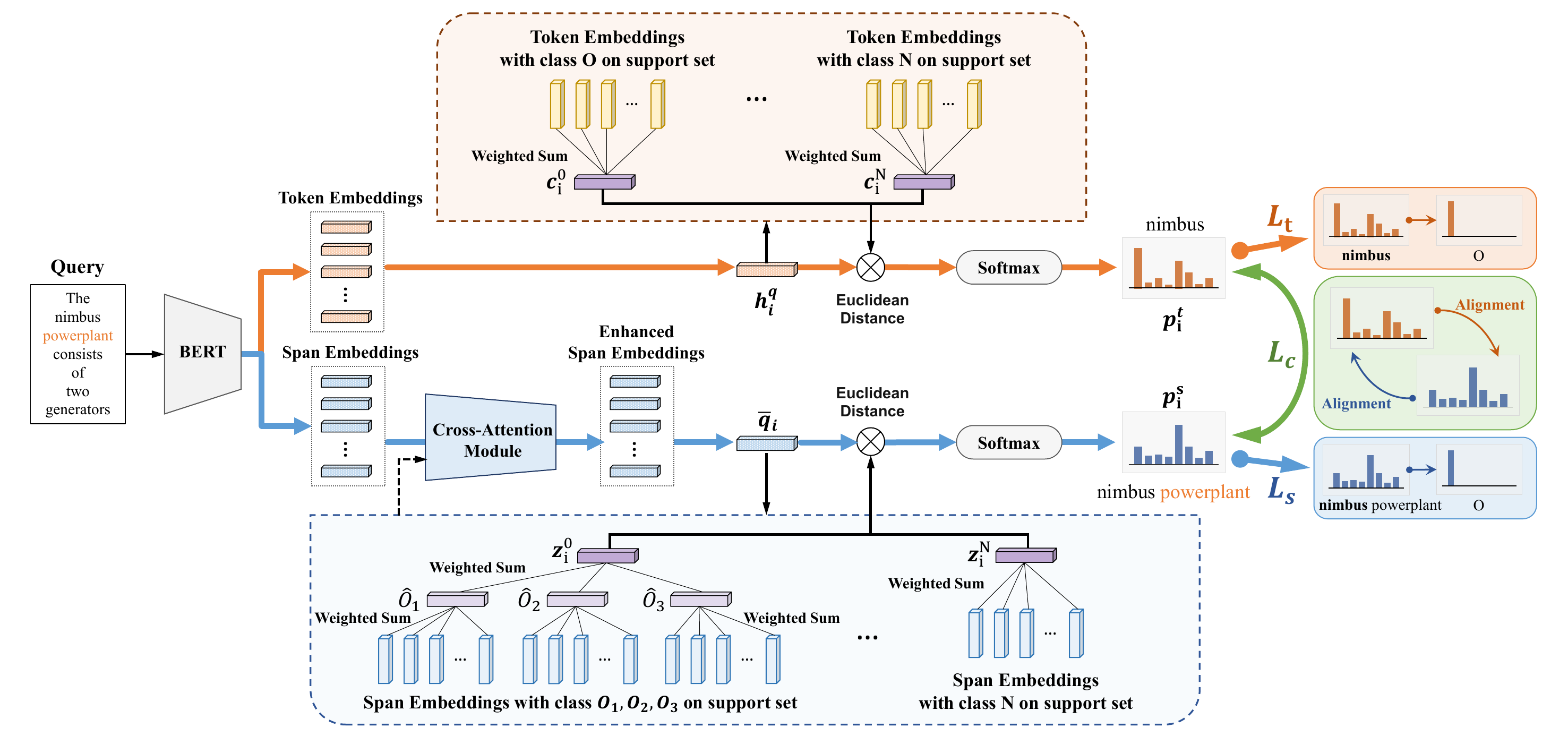}
\caption{The framework of our proposed model. The upper part corresponds to token-level network and the lower part corresponds to span-level network.} \label{fig:framework}
\end{figure*}

In this section, we present the details of our proposed model, as shown in Figure~\ref{fig:framework}.
Our model utilizes both token-level and span-level labels to mitigate the data scarcity problem in the few-shot setting from a unified perspective.
We first explain the encoder module of our model (Section~\ref{sec:encoder}).
Next we present the model architecture of token-level adaptive prototypical network (Section~\ref{sec:token}) and span-level adaptive prototypical network (Section~\ref{sec:span}).
We further propose a consistent loss to make the predictions of two networks consistent (Section~\ref{sec:consistent}).
Then we introduce final loss function and training flow of our model (Section~\ref{sec:training}).
Finally, we propose a consistent greedy inference algorithm to combine the outputs of two networks during the inference phase (Section~\ref{sec:inference}).

\subsection{Encoder Module} \label{sec:encoder}
We utilize BERT~\cite{DBLP:conf/naacl/DevlinCLT19} to encode sentences in the support and query set.
Given a sentence $\boldsymbol{x} = \{x_1, x_2, ..., x_n\}$, we insert a [CLS] and [SEP] token for each sentence to get the input of BERT and feed it into BERT to get its contextual representations $\mathbf{U} = \{\mathbf{u_0}, \mathbf{u_1}, \mathbf{u_2}, ..., \mathbf{u_n}, \mathbf{u_{n+1}}\} \in \mathbb{R}^{(n+2) \times d_1}$ as follows:
\begin{equation}
\mathbf{u_0}, \mathbf{u_1}, \mathbf{u_2}, ..., \mathbf{u_n}, \mathbf{u_{n+1}} = \textrm{BERT}(\textrm{[CLS]}, x_1, x_2, ..., x_n, \textrm{[SEP]})
\end{equation}

We further use a linear layer to obtain the token representation $\mathbf{H} = \{\mathbf{h_1}, \mathbf{h_2}, ..., \mathbf{h_n}\} \in \mathbb{R}^{n \times d}$ as the input to the token-level network as follows:
\begin{equation}
\mathbf{h_i} = \mathbf{W_t}\mathbf{u_i} + \mathbf{b_t}, i=1, 2, ..., n
\end{equation}
where $\mathbf{W_t} \in \mathbb{R}^{d \times d_1}$ and $\mathbf{b_t} \in \mathbb{R}^{d}$ are trainable parameters.
Since [CLS] and [SEP] tokens do not need to be classified, we do not feed them into the linear layer.

For the span-level network, we use boundary tokens (i.e., left and right boundary tokens) and a linear layer to obtain the span representation as the input to the span-level network.
Specifically, given a span $\mathbf{s}_{k} = \{x_i, ..., x_j\}$, we represent the span as:
\begin{equation}
\mathbf{s}_{k} = \mathbf{W_s}[\mathbf{u_i} \oplus \mathbf{u_j}] + \mathbf{b_s}
\end{equation}
where $\mathbf{W_s} \in \mathbb{R}^{d \times 2d_1}$ and $\mathbf{b_s} \in \mathbb{R}^{d}$ are trainable parameters, and $\oplus$ is the concatenation operation.
It is worth noting that we enumerate all spans in the sentence.

\subsection{Token-Level Adaptive Prototypical Network} \label{sec:token}
For the token-level network, we adopt an adaptive prototypical network to classify tokens.
General prototypical networks~\cite{snell2017prototypical} simply average all the tokens of the same class to get prototype.
However, the average operation does not reflect the importance of each token~\cite{DBLP:conf/aaai/GaoH0S19,DBLP:conf/emnlp/SunSZL19}.
Therefore, we propose to construct adaptive prototypes for each query token by considering the importance of each token.
Specifically,
\begin{equation}
\mathbf{\alpha_i^j} = \textrm{softmax}(\mathbf{H^s_j} \mathbf{h_i^q})
\end{equation}
\begin{equation}
\mathbf{c_i^j} = (\mathbf{H^s_j})^{\mathrm{T}} \mathbf{\alpha_i^j}
\end{equation}
where $\mathbf{h_i^q} \in \mathbb{R}^{d}$ is representation of query token $x_i$, $\mathbf{H^s_j} \in \mathbb{R}^{C_j \times d}$ is representation of all tokens of class $j$ in the support set, $C_j$ is the number of $j$-class tokens in support set, $\mathbf{\alpha_i^j} \in \mathbb{R}^{C_j}$ is attention weight, and $\mathbf{c_i^j} \in \mathbb{R}^{d}$ is adaptive prototype of class $j$ for query token $x_i$.
For clarity, we denote this attention aggregation as $\phi$, i.e., $\mathbf{c_i^j} = \phi(\mathbf{h_i^q}, \mathbf{H^s_j})$.

Then we utilize cross-entropy loss to optimize the network and the predicted probability distribution can be produced by a softmax function over Euclidean distances between all prototypes and query token $x_i$.
Specifically,
\begin{equation}
p_i^t(\hat{y}_i = j | x_i) = \frac{\textrm{exp}(-d(\mathbf{h_i^q},\mathbf{c_i^j}))} {\sum_{p}\textrm{exp}(-d(\mathbf{h_i^q},\mathbf{c_i^p}))}
\end{equation}
\begin{equation}
\mathcal{L}_t = - \sum_{i=1}^{n}\textrm{log} p_i^t(\hat{y}_i = y_i^*|x_i)
\end{equation}
where $p_i^t(\hat{y}_i = j | x_i)$ is predicted probability of $j$-class for token $x_i$, $d$ denotes Euclidean distance, and $y_i^*$ denotes the ground-truth label for token $x_i$.
It is worth noting that we use the IO (Inside and Outside) scheme for token-level network same as~\citet{DBLP:conf/acl/DingXCWHXZL21}, not BIESO (Begin, Inside, End, Singleton, and Outside) or BIO (Begin, Inside, and Outside) scheme.
For example, the label space of example in Table 1 is \{building-other, building-library, Outside\}, not \{begin-building-other, inside-building-other, end-building-other, singleton-building-other, begin-building-library, inside-building-library, end-building-library, singleton-building-library, outside\} or \{begin-building-other, inside-building-other, begin-building-library, inside-building-library, outside\}.

\subsection{Span-Level Adaptive Prototypical Network}\label{sec:span}
For the span-level network, we classify spans by adopting an adaptive prototypical network similar to the token-level adaptive prototypical network, the difference mainly lies in that we employ an extra cross-attention module and make more fine-grained division for class O.

\subsubsection{Cross-Attention Module}
To recognize a sample from novel class given a few labeled samples, a natural idea is to locate the most relevant regions in the pair of support and query samples~\cite{DBLP:conf/nips/HouCMSC19}.
Thus, we use a cross-attention to model the interaction between support set and query set to enhance the representation of support set and query set.
We denote the span representation in the support and query set as $\mathbf{S} \in \mathbb{R}^{S_s \times d}$ and $\mathbf{Q} \in \mathbb{R}^{S_q \times d}$, where $S_s$ and $S_q$ denote the number of spans in the support and query set.
Then we use a cross-attention to model the interaction: 
\begin{equation}
\hat{\mathbf{s}}_n = \phi(\mathbf{s_n}, \mathbf{Q})
\end{equation}
\begin{equation}
\hat{\mathbf{q}}_n = \phi(\mathbf{q_n}, \mathbf{S})
\end{equation}
where $\mathbf{s}_n \in \mathbb{R}^{d}$ and $\mathbf{q}_n \in \mathbb{R}^{d}$ denote the $n$-row of $\mathbf{S}$ and $\mathbf{Q}$, $\hat{\mathbf{s}}_n \in \mathbb{R}^{d}$ denotes the span representation after cross-attention in the support set, and $\hat{\mathbf{q}}_n \in \mathbb{R}^{d}$ denotes the span representation after cross-attention in the query set.
Finally, we use Transformer block~\cite{DBLP:conf/nips/VaswaniSPUJGKP17} to get the enhanced representation:
\begin{equation}
\bar{\mathbf{s}}_n = \textrm{LayerNorm}(\mathbf{s}_n+\textrm{FFN}(\hat{\mathbf{s}}_n))
\end{equation}
\begin{equation}
\bar{\mathbf{q}}_n = \textrm{LayerNorm}(\mathbf{q}_n+\textrm{FFN}(\hat{\mathbf{q}}_n))
\end{equation}
where $\bar{\mathbf{s}}_n \in \mathbb{R}^{d}$ and $\bar{\mathbf{q}}_n \in \mathbb{R}^{d}$ denote enhanced representation of support and query set, and FFN denotes a feed forward neural network.

\subsubsection{Fine-Grained Division of O}
Different from entity classes, non-entity class O has rich semantics and is hard to represent with a single prototype~\cite{DBLP:conf/acl/TongWX0L0L21,DBLP:conf/naacl/WangXLZCCS22}.
So we further divide class O into three sub-classes.
Since we use left and right boundary tokens to represent span, we divide sub-classes based on whether the span has the same left and right boundary as entity spans.
Specifically, given a sentence with $M_1$ entity spans $\{(l_i, r_i)\}_{i=1}^{M_1}$, where $l_i$ and $r_i$ are the left and right boundary tokens of the $i$-th span, other non-entity spans can be assigned with a sub-class $O_{sub}$ label as follows,
\begin{align*}
\begin{split}
O_{sub}= \left \{
\begin{array}{ll}
    O_1, & \exists i, l_o = l_i\\
    O_2, & \exists i, r_o = r_i \\
    O_3, & \textrm{others}
\end{array}
\right.
\end{split}
\end{align*}
where $O_1$ denotes the non-entity spans that have the same left boundary token as entity spans, $O_2$ denotes the non-entity spans that have the same right boundary token as entity spans, and $O_3$ denotes the non-entity spans do not have the same left and right boundary tokens as entity spans.
It is worth noting that for sentences with two or more entities, there may be some non-entity spans that have the same left boundary with one entity span and same right boundary with another entity span.
We treat these non-entity spans as $O_1$.

\begin{algorithm}[t]
\caption{The Training Flow of Consistent Dual Adaptive Prototypical Network}\label{alg:train}
\begin{algorithmic}
\Require The few-shot sequence labeling training set $\mathcal{D}_{tr}$
\Ensure A well-trained few-shot sequence labeling model
\ForAll{iteration = 1, $\cdots$, MaxIter}
\State Randomly sample a mini-batch data ($\mathcal{S}_{train}$, $\mathcal{Q}_{train}$) from $\mathcal{D}_{tr}$
\State $\rhd$ Base Encoder
\State $\mathbf{H}^\mathbf{s}, \mathbf{S} \gets$ BERT($\mathcal{S}_{train}$)
\State $\mathbf{H}^\mathbf{q}, \mathbf{Q} \gets$ BERT($\mathcal{Q}_{train}$)
\ForEach{$x$ in $\mathcal{Q}_{train}$}
\State $\rhd$ Token-Level Network
\State Generate token-level prototype: $\mathbf{c_i^j} = \phi(\mathbf{h_i^q}, \mathbf{H^s_j})$
\State Calculate token-level loss: $\mathcal{L}_t = - \sum_{i=1}^{n}\text{log} \frac{\textrm{exp}(-d(\mathbf{h_i^q},\mathbf{c_i^*}))} {\sum_{p}\textrm{exp}(-d(\mathbf{h_i^q},\mathbf{c_i^p}))}$
\State $\rhd$ Span-Level Network
\State Enhance representation: $\bar{\mathbf{S}}, \bar{\mathbf{Q}} \gets$ Cross-attention module ($\mathbf{S}$, $\mathbf{Q}$)
\State Generate span-level prototype: $\mathbf{z_i^j} = \phi(\bar{\mathbf{q}}_i, \bar{\mathbf{S}}_j)$
\State Calculate span-level loss: $\mathcal{L}_s = - \sum_{i=1}^{S_q}\text{log} \frac{\textrm{exp}(-d(\bar{\textbf{q}}_i,\mathbf{z_i^*}))} {\sum_{p}\textrm{exp}(-d(\bar{\textbf{q}}_i,\mathbf{z_i^p}))}$
\State $\rhd$ Consistent Loss Calculation
\State Generate two token-level distributions $l_t(x_i)$ and $l_s(x_i)$ from two networks
\State Calculate loss: $\mathcal{L}_{c}$ = $\sum_{i=1}^n [D_{KL}\left(\sigma(l_t(x_i)/T)||\sigma(l_s(x_i))\right)$ + $D_{KL}\left(\sigma(l_s(x_i)/T)||\sigma(l_t(x_i))\right)]$
\State $\rhd$ Optimization
\State Obtain total loss $\mathcal{L} = \lambda \mathcal{L}_{t} + \beta \mathcal{L}_s  + \gamma \mathcal{L}_c$ and update model
\EndFor
\EndFor
\State \Return The well-trained model
\end{algorithmic}
\end{algorithm}

\subsubsection{Adaptive Prototypical Network}
We also construct adaptive prototypes for each query span, similar to token-level adaptive prototypical network.
Since we divide class O into three sub-classes, we first adaptively construct prototype of each sub-class and then adaptively aggregate three sub-classes to get the prototype of class O.
For entity classes, we can directly get the adaptive prototypes.
Specifically,
\begin{equation}
\hat{\textbf{o}}_k = \phi(\bar{\textbf{q}}_i, \textbf{O}_k),  k=1, 2, 3
\end{equation}
\begin{equation}
\textbf{z}_i^0 = \phi(\bar{\textbf{q}}_i, \left[
\begin{array}{ccc}
\hat{\textbf{o}}_1^{\mathrm{T}}\\
\hat{\textbf{o}}_2^{\mathrm{T}}\\
\hat{\textbf{o}}_3^{\mathrm{T}}\\
\end{array}
\right])
\end{equation}
\begin{equation}
\textbf{z}_i^k = \phi(\bar{\textbf{q}}_i, \bar{\textbf{S}}_k) , k=1, 2, ..., N
\end{equation}
where $\hat{\textbf{o}}_k \in \mathbb{R}^{d}$ denotes the adaptive prototype of sub-class $O_k$ for query span $q_i$, $\textbf{O}_k \in \mathbb{R}^{o_{k}\times d}$ denotes the representation of all spans in the support set with sub-class $O_k$, $o_k$ denotes the number of spans with sub-class $O_k$ in the support set, $\textbf{z}_i^0 \in \mathbb{R}^{d}$ denotes the adaptive prototype of class O for query span $q_i$, $\bar{\textbf{S}}_k \in \mathbb{R}^{m_{k}\times d}$ is representation of all spans of class $k$ in the support set, $m_k$ denotes the number of $k$-class spans in the support set, and $\textbf{z}_i^k \in \mathbb{R}^{d}$ denotes the adaptive prototype of class $k$ for query span $q_i$.

After getting all prototypes $\textbf{Z}_i = (\textbf{z}_i^0, \textbf{z}_i^1, ..., \textbf{z}_i^N)$, we also use softmax function over Euclidean distances between all prototypes and query span $q_i$ to get the predicted probability distribution, and cross-entropy loss to optimize the network.
\begin{equation}
p_i^s(\hat{y}_i = j | q_i) = \frac{\textrm{exp}(-d(\bar{\textbf{q}}_i,\mathbf{z_i^j}))} {\sum_{p}\textrm{exp}(-d(\bar{\textbf{q}}_i,\mathbf{z_i^p}))}
\end{equation}
\begin{equation}
\mathcal{L}_s = - \sum_{i=1}^{S_q}\textrm{log} p_i^s(\hat{y}_i = \bar{y}_i^*|q_i)
\end{equation}
where $p_i^s(\hat{y}_i = j | q_i)$ is predicted probability of $j$-class for span $q_i$, $\bar{y}_i^*$ denotes the ground-truth label of span $q_i$, and $S_q$ is the number of spans in the sentence.

\subsection{Consistent Loss} \label{sec:consistent}
Intuitively, the token-level network and span-level network should make consistent predictions in a well-trained model.
Therefore, we further propose a consistent loss to make the predicted probability distributions of two networks consistent.
It contains two steps, where the first step constructs two token-level probability distributions for each token from two networks and the second step measures the difference between two probability distributions.

For each token, we can obtain the token-level probability distribution directly from the token-level network.
Then, among all spans containing this token, we select the span with the maximum predicted probability and take its probability distribution as the token-level probability distribution from span-level network.
Finally, we use bidirectional Kullback-Leibler divergence with temperature as consistent loss to allow the two networks to learn from each other and to measure the difference between the two distributions.
Specifically,
 \begin{equation}
 \begin{aligned}
    \mathcal{L}_{c} = \sum_{i=1}^{n} [D_{KL}\left(\sigma(l_t(x_i)/T)||\sigma(l_s(x_i))\right) + 
    D_{KL}\left(\sigma(l_s(x_i)/T)||\sigma(l_t(x_i))\right)]
\end{aligned}
\end{equation}
where $l_t(x_i)$ and $l_s(x_i)$ are the predicted logits of token-level and span-level networks for token $x_i$, $\sigma$ denotes softmax function, and $T$ is the temperature to adjust the sharpness of the probability distribution.

\subsection{Training}  \label{sec:training}
Finally, we combine the above three loss functions to form the final loss function to jointly train our model as follows:
\begin{equation}
\mathcal{L} = \lambda \mathcal{L}_{t} + \beta \mathcal{L}_s  + \gamma \mathcal{L}_c
\end{equation}
where $\lambda$, $\beta$, $\gamma$ are hyper-parameters.

In summary, there are three steps to train our model, i.e., calculate token-level loss, calculate span-level loss, and calculate consistent loss. 
The whole training flow is illustrated in Figure~\ref{fig:framework} and Algorithm~\ref{alg:train}.

\begin{figure}[t]
\centering
\includegraphics[width=0.7\columnwidth]{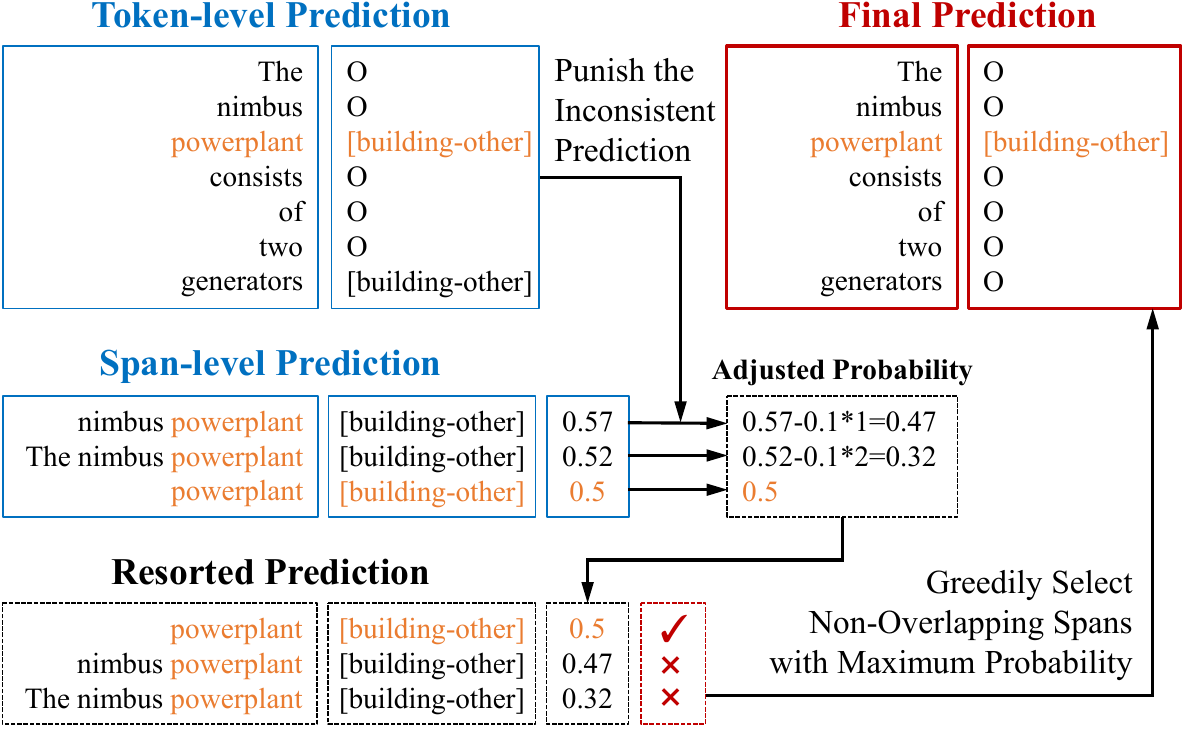}
\caption{The process of consistent greedy inference algorithm. The algorithm first punishes the inconsistent prediction and then greedily selects non-overlapping spans with maximum probability based on adjusted probability.} \label{fig:inference}
\end{figure}

\subsection{Consistent Greedy Inference Algorithm} \label{sec:inference}
Intuitively, we consider extracted entities and slots unreliable if the two networks predict different labels.
So we propose a consistent greedy algorithm to combine the outputs of the two networks.
As shown in Figure~\ref{fig:inference} and Algorithm~\ref{alg:infer}, we first adjust the predicted probabilities of spans and then use a greedy algorithm to gradually select non-overlapping spans with the maximum probability based on the adjusted probability.

First, we consider a span to be reliable if all token-level predictions from token-level network in the span agree with the span-level prediction from span-level network.
For example, as shown in Figure~\ref{fig:inference}, the prediction of span ``powerplant'' is consistent, and the prediction of ``nimbus powerplant'' and ``the nimbus powerplant'' is inconsistent.
We punish the probabilities of inconsistent spans by the following operation:
\begin{equation}
\bar{y}_{i} = \hat{y}_{i} - \delta * \textrm{count}_{i}
\end{equation}
where $\bar{y}_{i}$ and $\hat{y}_{i}$ is the predicted probability of span $q_{i}$ after and before consistent adjustment, $\delta$ is a hyper-parameter to control the degree of penalty, and $\textrm{count}_{i}$ is the number of inconsistent tokens.
Then, we use a greedy algorithm to gradually select non-overlapping spans with the maximum probability based on the adjusted probability.
Specifically, at each step of the greedy algorithm, we select the span with the maximum probability and discard the overlapping span.
For example, we select span ``powerplant'' and discard ``nimbus powerplant'' and ``the nimbus powerplant''.
The whole inference flow is illustrated in Algorithm~\ref{alg:infer}.

\begin{algorithm}[t]
\caption{The Process of Consistent Greedy Inference Algorithm}\label{alg:infer}
\begin{algorithmic}
\Require The few-shot sequence labeling test set $\mathcal{D}_{te}$, a well-trained model, the number of episodes in test set $n$
\Ensure The extracted entities or slots set S
\State S = $\emptyset$
\ForAll{i = 1, $\cdots$, $n$}
\State \textbf{\emph{\# Initialization: get data and predictions from two networks:}}
\State B, C = $\emptyset$, $\emptyset$
\State Get i-th episode data ($\mathcal{S}_{test}$, $\mathcal{Q}_{test}$) from $\mathcal{D}_{te}$
\State Get token-level prediction $\{\hat{y}_1, \hat{y}_2, ..., \hat{y}_n\}$ according to Eq. (6)
\State Get span-level prediction A = $\{(s_j,\hat{y}_j)\}_{j=1}^{\hat{M}}$ according to Eq. (15)
\State \textbf{\emph{\# First step: punish the inconsistent prediction:}}
\ForAll{k = 1, $\cdots$, $\hat{M}$}
\State Get the number of inconsistent tokens $\textrm{count}_{k}$ in span $s_k$
\State $\bar{y}_k \gets \hat{y}_{k} - \delta * \textrm{count}_{k}$
\State A $\gets$ A $\setminus$ $\{(s_k,\hat{y}_k)\}$
\State B $\gets$ B $\cup$ ($s_k$,$\bar{y}_k$)
\EndFor
\State \textbf{\emph{\# Second step: greedy algorithm to select spans:}}
\While{B $\neq \emptyset$}
        \State Select span $s_k$ with maximum probability in B
        \State C $\gets$ C $\cup$ $s_k$
        \State Get span set D that overlap with span $s_k$ in B
        \State B $\gets$ B $\setminus$ D
\EndWhile
\State S $\gets$ S $\cup$ C
\EndFor
\State \Return The extracted entities or slots set S
\end{algorithmic}
\end{algorithm}

\section{Experiments} \label{sec:exp}
In this section, we first introduce our experimental details, including the datasets, evaluation metrics, implementation details, and baselines.
Then, we report the experimental results on three datasets to answer the following research questions:
\begin{itemize}
\item \textbf{RQ1}: Whether our proposed model outperforms existing few-shot sequence labeling methods?
\item \textbf{RQ2}: How does each of the components of our model contribute to the final performance?
\item \textbf{RQ3}: What is effect of different inference algorithms on the performance of our method?
\item \textbf{RQ4}: What is effect of the different consistent losses on the performance of our method?
\item \textbf{RQ5}: How do the hyper-parameters influence the performance of our method?
\end{itemize}
Thereafter, we also explore the effectiveness of our O division strategy, evaluate the efficiency of our model, and conduct error analysis and case study to further analyze our model.

\begin{table}
\centering
\caption{Statistics of three original datasets.}\label{statis}
\begin{tabular}{lccc}
\toprule
\textbf{Datasets}& \textbf{\# Sentences}& \textbf{\# Classes}&\textbf{Domain}\\
\midrule
\textbf{FewNERD} & 188.2K & 66 & Wikipedia\\
\textbf{SNIPS} & 14.5K    & 60 & Dialog\\
\textbf{Cross} & 189.4K   & 43 & News, Wiki, Social, and Mixed\\
\bottomrule
\end{tabular}
\end{table}

\subsection{Datasets}
We conduct experiments on three benchmark datasets: FewNERD~\cite{DBLP:conf/acl/DingXCWHXZL21}, SNIPS~\cite{snips}, and Cross.
FewNERD and Cross are NER datasets, and SNIPS is a slot tagging dataset.
The statistics of the three original datasets are shown in Table~\ref{statis}.
The detailed description of the three datasets is as follows:

FewNERD dataset with a hierarchy of 8 coarse-grained entity types and 66 fine-grained entity types, and has two few-shot settings: Intra and Inter.
The two settings divide the dataset using coarse-grained and fine-grained entity types, respectively.
For \textbf{Intra} setting, all entities in the training set, development set, and test set belong to different coarse-grained types.
For \textbf{Inter} setting, all entities in the training set, development set, and test set belong to different fine-grained types.
It is worth noting that in order to better meet the dense entities, i.e., a sentence may have multiple entity types, \citet{DBLP:conf/acl/DingXCWHXZL21} adopt the $N$-way $K$\textasciitilde$2K$ shot sampling method to construct FewNERD dataset (each class in the support set has $K$\textasciitilde$2K$ entities).
Both Intra and Inter setting have 4 settings: 5-way 1\textasciitilde2 shot, 5-way 5\textasciitilde10 shot, 10-way 1\textasciitilde2 shot, and 10-way 5\textasciitilde10 shot.
We use the public sampled dataset\footnote{It is worth noting that we use the latest version of FewNERD dataset from \url{https://github.com/thunlp/Few-NERD}, which corresponds to the results reported in \url{https://arxiv.org/pdf/2105.07464v6.pdf}.} released by~\citet{DBLP:conf/acl/DingXCWHXZL21} which contains 20,000 episodes for training, 1,000 episodes for validation, and 5,000 episodes for testing.

SNIPS is a slot tagging dataset and contains 7 domains.
The domains are Weather (We), Music (Mu), Play List (Pl), Book (Bo), Search Screen (Se), Restaurant (Re), and Creative Work (Cr).
\citet{DBLP:conf/acl/HouCLZLLL20} randomly sample a domain for validation, then randomly sample a different domain for testing, and the other domains are used for training.
In each episode, all classes have $K$ shot samples in the support set and the number of classes ($N$) is not fixed.
Each domain of SNIPS dataset has two settings: 1-shot and 5-shot.
We use the public sampled dataset\footnote{\url{https://github.com/AtmaHou/FewShotTagging}} provided by~\citet{DBLP:conf/acl/HouCLZLLL20} for a fair comparison.

Cross is a NER dataset, consisting of four datasets CoNLL-2003~\cite{DBLP:conf/conll/SangM03}, GUM~\cite{DBLP:journals/lre/Zeldes17}, WNUT-2018~\cite{DBLP:conf/aclnut/DerczynskiNEL17}, and Ontonotes~\cite{DBLP:conf/conll/PradhanMXNBUZZ13}.
The four datasets are from News, Wiki, Social, and Mixed domains respectively.
Same as the SNIPS dataset,~\citet{DBLP:conf/acl/HouCLZLLL20} use two datasets for training, one dataset for validation, and one dataset for testing.
Each domain of Cross dataset also has two settings: 1-shot and 5-shot and we also use the public sampled dataset\footnote{\url{https://github.com/AtmaHou/FewShotTagging}} provided by~\citet{DBLP:conf/acl/HouCLZLLL20} for a fair comparison.

\begin{table*}[t]\small \setlength{\tabcolsep}{0.001pt}
  \centering
      \caption{The performance (F1 score with standard deviations) on FewNERD dataset. $\dag$ denotes that we reproduce the results using the authors' code\protect\footnotemark\ and do not fine-tune model during meta-testing phase for a fair comparison. We report the average result of 5 runs.}     \label{fewnerd}
    \begin{tabular}{cccccccccccccccc}
    \toprule
    \multirow{3}{*}{\textbf{Models}}&\multicolumn{5}{c}{\textbf{Intra}}&\multicolumn{5}{c}{\textbf{Inter}}\\
    \cmidrule(lr){2-6} \cmidrule(lr){7-11} \cmidrule(lr){8-10}&\multicolumn{2}{c}{\textbf{1\textasciitilde2 shot}}&\multicolumn{2}{c}{\textbf{5\textasciitilde10 shot}}&\multirow{2}{*}{\textbf{Avg.}}&\multicolumn{2}{c}{\textbf{1\textasciitilde2 shot}}&\multicolumn{2}{c}{\textbf{5\textasciitilde10 shot}}&\multirow{2}{*}{\textbf{Avg.}}\\
    \cmidrule(lr){2-3} \cmidrule(lr){4-5} \cmidrule(lr){7-8} \cmidrule(lr){9-10}&\textbf{5 way}&\textbf{10 way}&\textbf{5 way}&\textbf{10 way}&&\textbf{5 way}&\textbf{10 way}&\textbf{5 way}&\textbf{10 way}\\
    \midrule
    Proto           
    & 20.76\tiny{$\pm$0.84}   & 15.04\tiny{$\pm$0.44} & 42.54\tiny{$\pm$0.94} & 35.40\tiny{$\pm$0.13} & 28.44\tiny{$\pm$0.59}
            &  38.83\tiny{$\pm$1.49} &  32.45\tiny{$\pm$0.79}  & 58.79\tiny{$\pm$0.44}  & 52.92\tiny{$\pm$0.37} & 45.75\tiny{$\pm$0.77} \\
    NNShot& 25.78\tiny{$\pm$0.91}   & 18.27\tiny{$\pm$0.41} &  36.18\tiny{$\pm$0.79} & 27.38\tiny{$\pm$0.53} & 26.90\tiny{$\pm$0.66}& 47.24\tiny{$\pm$1.00} &  38.87\tiny{$\pm$0.21} & 55.64\tiny{$\pm$0.63}  & 49.57\tiny{$\pm$2.73}  & 47.83\tiny{$\pm$1.14}\\
    StructShot & 30.21\tiny{$\pm$0.90}   & 21.03\tiny{$\pm$1.13} &  38.00\tiny{$\pm$1.29} & 26.42\tiny{$\pm$0.60} & 28.92\tiny{$\pm$0.98} & 51.88\tiny{$\pm$0.69} &  43.34\tiny{$\pm$0.10}  & 57.32\tiny{$\pm$0.63}  & 49.57\tiny{$\pm$3.08} & 50.53\tiny{$\pm$1.13}\\
    De-MAML$^{\dag}$&37.49\tiny{$\pm$2.10}&27.36\tiny{$\pm$2.70}&41.26\tiny{$\pm$1.34}&35.31\tiny{$\pm$5.53}&35.36\tiny{$\pm$2.92}&61.63\tiny{$\pm$1.11}&49.24\tiny{$\pm$5.10}&54.09\tiny{$\pm$2.24}&53.20\tiny{$\pm$3.44}&54.54\tiny{$\pm$2.97}\\
    ESD& 36.08\tiny{$\pm$1.60}   & 30.00\tiny{$\pm$0.70} &  52.14\tiny{$\pm$1.50} & 42.15\tiny{$\pm$2.60} & 40.09\tiny{$\pm$1.60} &   59.29\tiny{$\pm$1.25}&  52.16\tiny{$\pm$0.79}  & 69.06\tiny{$\pm$0.80}  & 64.00\tiny{$\pm$0.43} & 61.13\tiny{$\pm$0.82}\\
            Ours& \textbf{41.21\tiny{$\pm$0.70}}& \textbf{35.36\tiny{$\pm$0.91}} & \textbf{52.67\tiny{$\pm$0.49}}&\textbf{45.83\tiny{$\pm$0.81}} & \textbf{43.77\tiny{$\pm$0.73}} & \textbf{62.79\tiny{$\pm$0.29}} &\textbf{56.23\tiny{$\pm$0.76}} & \textbf{69.83\tiny{$\pm$0.40}} &\textbf{65.99\tiny{$\pm$0.40}} & \textbf{63.71\tiny{$\pm$0.46}}\\
    \bottomrule
    \end{tabular}
\end{table*}
\footnotetext{\url{https://github.com/microsoft/vert-papers/tree/master/papers/DecomposedMetaNER}.}

\subsection{Evaluation Metrics} 
For the FewNERD dataset, we report the micro-F1 over all test episodes following~\citet{DBLP:conf/acl/DingXCWHXZL21}.
Specifically,
\begin{equation}
\textrm{F1} = \frac{2 \times \textrm{P}_{all} \times \textrm{R}_{all}}{\textrm{P}_{all}+\textrm{R}_{all}}
\end{equation}
where $\textrm{P}_{all}$ denotes the precision calculated on all test episodes and $\textrm{R}_{all}$ denotes the recall calculated on all test episodes.
For the SNIPS and Cross dataset, we first calculate micro-F1 for each test episode and then report the average F1 for all test episodes as the final result following~\citet{DBLP:conf/acl/HouCLLC22}.
Specifically,
\begin{equation}
\textrm{F1} = \frac{1}{n}\sum_{i=1}^n \textrm{F1}_i
\end{equation}
where n is the number of episodes and $\textrm{F1}_i$ denotes the F1 of i-th episode.

\subsection{Implementation Details} 
We use BERT-base-uncased~\cite{DBLP:conf/naacl/DevlinCLT19} as our encoder and use AdamW~\cite{DBLP:conf/iclr/LoshchilovH19} optimizer.
For the FewNERD dataset, the learning rate is 2e-5 for BERT and 5e-4 for other parameters, and the batch size is set to 2.
For the SNIPS dataset, we use grid search to find the best hyper-parameters.
The learning rate of BERT is selected from \{5e-6, 1e-5, 2e-5, 3e-5, 5e-5\} and other parameters are selected from \{5e-5, 1e-4, 3e-4, 5e-4\}, and the batch size is selected from \{2, 4\}.
We schedule the learning rate that the first 1000 training steps is a linear warmup phrase and then the rest is a linear decay phrase.
$\lambda$ and $\gamma$ are selected from \{0.02, 0.05, 0.1, 0.2, 0.3\} through grid search, $\beta$ is set to 1, and $\delta$ is selected from \{0.01, 0.02, 0.05, 0.1\}.
The maximum span length is set to 8 for inference.

\subsection{Baselines}
To evaluate the effectiveness of our method, we compare our method with the following baselines:
\begin{itemize}
\item \textbf{Proto}~\cite{DBLP:conf/sac/FritzlerLK19,DBLP:conf/acl/DingXCWHXZL21} represents each class by the mean of its token representation and uses Euclidean distance to predict query labels.
\item \textbf{NNShot}~\cite{DBLP:conf/emnlp/YangK20,DBLP:conf/acl/DingXCWHXZL21} is similar to \textbf{Proto}, but uses nearest neighbor classification to predict query labels.
\item \textbf{StructShot}~\cite{DBLP:conf/emnlp/YangK20,DBLP:conf/acl/DingXCWHXZL21} is similar to \textbf{NNShot} and further uses a Viterbi decoder during the inference phase to model label dependency.
\item \textbf{TransferBERT}~\cite{DBLP:conf/acl/HouCLZLLL20} is a domain transfer model which pretrains model on source domains and fine-tunes on the target domain.
\item \textbf{Matching}~\cite{vinyals2016matching,DBLP:conf/acl/HouCLZLLL20} is similar to \textbf{Proto} and uses the matching networks~\cite{vinyals2016matching} to classification instead of prototypical networks.
\item \textbf{L-TapNet+CDT}~\cite{DBLP:conf/acl/HouCLZLLL20} uses label name to extend task-adaptive projection networks~\cite{DBLP:conf/icml/YoonSM19} and further utilizes collapsed dependency transfer to capture the dependencies between labels.
\item \textbf{Retriever}~\cite{DBLP:conf/naacl/YuHZDPL21} is a span-level retrieval method that learns similar contextualized representations for spans with the same label via a novel batch-softmax objective. 
\item \textbf{ConVEx}~\cite{DBLP:conf/naacl/HendersonV21} first pre-trains through a novel pairwise cloze task using Reddit data and then directly fine-tunes for slot tagging task.
\item \textbf{MRC}~\cite{DBLP:conf/acl/MaYLZ21} enumerates all the slots to extract the answer from the sentence in machine reading comprehension framework.
\item \textbf{Inverse}~\cite{DBLP:conf/acl/HouCLLC22} reversely predicts slot values given slot types prompt and further proposes iterative prediction strategy to consider the relations between
different slot types.
\item \textbf{De-MAML}~\cite{DBLP:conf/acl/MaJWZL22} proposes a two-step MAML-enhanced model to first detect span and then classify span.
\item \textbf{ESD}~\cite{DBLP:conf/naacl/WangXLZCCS22} proposes inter-span and cross-span attention to enhance the span representations, and uses instance span attention to construct prototype.
\end{itemize}

\begin{table*}[t] \setlength{\tabcolsep}{1pt}
\centering
\caption{The performance (F1 scores with standard deviations) on $7$ domains of SNIPS. `unk' denotes methods that do not report standard deviations in their paper. Baselines of 1-shot and 5-shot settings are different since ConVEx, Retriever, and Inverse do not report the 1-shot results in their paper. We report the average result of 10 runs. Best result is \textbf{bold} and second best one is \underline{underlined}.} \label{tab:snips}
    \begin{tabular}{p{0.2cm}lcccccccccccc}
    \toprule
    &  \textbf{Models} & We & Mu & Pl & Bo & Se & Re & Cr & \textbf{Avg.} \\ 
      \midrule
     \multirow{7}{*}{\rotatebox{90}{\textbf{\textsc{1-shot}}} }   
     & TransferBERT   & {55.82\tiny{$\pm$2.75}} & {38.01\tiny{$\pm$1.74}} & {45.65\tiny{$\pm$2.02}} & {31.63\tiny{$\pm$5.32}} & {21.96\tiny{$\pm$3.98}} & {41.79\tiny{$\pm$3.81}} & {38.53\tiny{$\pm$7.42}} & {39.06\tiny{$\pm$3.86}}  \\
     & Matching  & {21.74\tiny{$\pm$4.60}} & {10.68\tiny{$\pm$1.07}} & {39.71\tiny{$\pm$1.81}} & {58.15\tiny{$\pm$0.68}} & {24.21\tiny{$\pm$1.20}} & {32.88\tiny{$\pm$0.64}} & {{69.66}\tiny{$\pm$1.68}} & {36.72\tiny{$\pm$1.67}}   \\
     & Proto   & {46.72\tiny{$\pm$1.03}} & {40.07\tiny{$\pm$0.48}} & {50.78\tiny{$\pm$2.09}} & {68.73\tiny{$\pm$1.87}} & {60.81\tiny{$\pm$1.70}} & {55.58\tiny{$\pm$3.56}} & {67.67\tiny{$\pm$1.16}} & {55.77\tiny{$\pm$1.70}}   \\
     &MRC & - & - & - & - & - & - & - & 69.3\tiny{(unk)} \\
     & L-TapNet+CDT & {{71.53}\tiny{$\pm$4.04}} & \textbf{{60.56}\tiny{$\pm$0.77}} & {{66.27}\tiny{$\pm$2.71}} & \textbf{84.54\tiny{$\pm$1.08}} &  \textbf{76.27\tiny{$\pm$1.72}} & {{70.79}\tiny{$\pm$1.60}} & {62.89\tiny{$\pm$1.88}} & {{70.41}\tiny{$\pm$1.97}} \\ 
     & ESD & \textbf{78.25\tiny{$\pm$1.50}} & {54.74\tiny{$\pm$1.02}} & \underline{71.15\tiny{$\pm$1.55}} & 71.45\tiny{$\pm$1.38} & 67.85\tiny{$\pm$0.75}& \underline{71.52\tiny{$\pm$0.98}}  &  \textbf{78.14\tiny{$\pm$1.46}} & \underline{70.44\tiny{$\pm$1.23}} \\
     &Ours& \underline{73.56\tiny{$\pm$1.58}}  & \underline{58.40\tiny{$\pm$1.42}} & \textbf{74.20\tiny{$\pm$1.24}} & \underline{76.07\tiny{$\pm$1.71}} &\underline{70.64\tiny{$\pm$1.49}} &\textbf{72.73\tiny{$\pm$1.08}} & \underline{75.32\tiny{$\pm$1.09}} &\textbf{71.56}\tiny{$\pm$1.37} \\
      \midrule
    
    \multirow{10}{*}{\rotatebox{90}{ \textbf{\textsc{5-shot}} }} 
     & TransferBERT& {59.41\tiny{$\pm$0.30}} & {42.00\tiny{$\pm$2.83}} & {46.07\tiny{$\pm$4.32}} & {20.74\tiny{$\pm$3.36}} & {28.20\tiny{$\pm$0.29}} & {67.75\tiny{$\pm$1.28}} & {58.61\tiny{$\pm$3.67}} & {46.11}\tiny{$\pm$2.29}  \\
     & Matching & {36.67\tiny{$\pm$3.64}} & {33.67\tiny{$\pm$6.12}} & {52.60\tiny{$\pm$2.84}} & {69.09\tiny{$\pm$2.36}} & {38.42\tiny{$\pm$4.06}} & {33.28\tiny{$\pm$2.99}} & {72.10\tiny{$\pm$1.48}} & {47.98}\tiny{$\pm$3.36}  \\
     & Proto 
        & {67.82\tiny{$\pm$4.11}} & {55.99\tiny{$\pm$2.24}} & {46.02\tiny{$\pm$3.19}} & {72.17\tiny{$\pm$1.75}} & {73.59\tiny{$\pm$1.60}} & {60.18\tiny{$\pm$6.96}} & {66.89\tiny{$\pm$2.88}} & {63.24}\tiny{$\pm$3.25}  \\
     & Retriever 
        & 82.95\tiny{(unk)}   & 61.74\tiny{(unk)}   & 71.75\tiny{(unk)}   & 81.65\tiny{(unk)}   & 73.10\tiny{(unk)}   & 79.54\tiny{(unk)}   & 51.35\tiny{(unk)}   & 71.72\tiny{(unk)}    \\
    & L-TapNet+CDT
        & {71.64\tiny{$\pm$3.62}} & {67.16\tiny{$\pm$2.97}} & {{75.88}\tiny{$\pm$1.51}} & 84.38\tiny{$\pm$2.81} & \underline{82.58\tiny{$\pm$2.12}} & {{70.05}\tiny{$\pm$1.61}} & {{73.41}\tiny{$\pm$2.61}} & {75.01}\tiny{$\pm$2.46}  \\
    &Inverse & 70.63\tiny{(unk)} & 71.97\tiny{(unk)} & 78.73\tiny{(unk)} & \textbf{87.34\tiny{(unk)}} & 81.95\tiny{(unk)} & 72.07\tiny{(unk)} & 74.44\tiny{(unk)} & 76.73\tiny{(unk)} \\
    & ConVEx  & 71.5\tiny{(unk)} & \textbf{77.6\tiny{(unk)}} & 79.0\tiny{(unk)} & 84.5\tiny{(unk)} & \textbf{84.0\tiny{(unk)}} & 73.8\tiny{(unk)} & 67.4\tiny{(unk)} & 76.8\tiny{(unk)}  \\
     & MRC  & \textbf{89.39\tiny{(unk)}} & \underline{75.11\tiny{(unk)}} & 77.18\tiny{(unk)} & 84.16\tiny{(unk)} & 73.53\tiny{(unk)} & \textbf{82.29\tiny{(unk)}} & 72.51\tiny{(unk)} & 79.17\tiny{(unk)}  \\
     & ESD
        & 84.50\tiny{$\pm$1.06}& 66.61\tiny{$\pm$2.00}& \underline{79.69\tiny{$\pm$1.35}} & 82.57\tiny{$\pm$1.37} & 82.22\tiny{$\pm$0.81} & 80.44\tiny{$\pm$0.80}& \textbf{81.13\tiny{$\pm$1.84}} & \underline{79.59\tiny{$\pm$1.32}}\\
    &Ours&\underline{86.24\tiny{$\pm$1.04}} & 67.66\tiny{$\pm$1.01}& \textbf{82.02\tiny{$\pm$1.05}}& \underline{86.47\tiny{$\pm$0.92}}& 82.45\tiny{$\pm$1.09}& \underline{81.38\tiny{$\pm$0.55}}& \underline{80.55\tiny{$\pm$1.16}} &\textbf{80.97}\tiny{$\pm$0.97}\\
     \bottomrule
    \end{tabular}
\end{table*}

\begin{table*}[t]\small \setlength{\tabcolsep}{0.0001pt}
  \centering
      \caption{The performance (F1 score with standard deviations) on the Cross Dataset. $\dag$ denotes that we reproduce the results using the authors' code and do not fine-tune model during meta-testing phase for a fair comparison. We report the average result of 10 runs.}  \label{cross-dataset}
    \resizebox{\linewidth}{!}{
    \begin{tabular}{cccccccccccccccc}
    \toprule
    \multirow{2}{*}{\textbf{Models}}&\multicolumn{5}{c}{\textbf{1-shot}}&\multicolumn{5}{c}{\textbf{5-shot}}\\
    \cmidrule(lr){2-6} \cmidrule(lr){7-11} & \textbf{News}& \textbf{Wiki} & \textbf{Social} & \textbf{Mixed} & \textbf{Avg.} & \textbf{News}& \textbf{Wiki} & \textbf{Social} & \textbf{Mixed} & \textbf{Avg.} \\
    \midrule
    TransferBERT           
    & 4.75\tiny{$\pm$1.42}   & 0.57\tiny{$\pm$0.32} & 2.71\tiny{$\pm$0.72} & 3.46\tiny{$\pm$0.54} & 2.87\tiny{$\pm$0.75}& 15.36\tiny{$\pm$2.81}
            &  3.62\tiny{$\pm$0.57} &  11.08\tiny{$\pm$0.57}  & 35.49\tiny{$\pm$7.60}  & 16.39\tiny{$\pm$2.89}\\
    Matching           
    & 19.50\tiny{$\pm$0.35}   & 4.73\tiny{$\pm$0.16} & 17.23\tiny{$\pm$2.75} & 15.06\tiny{$\pm$1.61} & 14.13\tiny{$\pm$1.22}& 19.85\tiny{$\pm$0.74}
            &  5.58\tiny{$\pm$0.23} &  6.61\tiny{$\pm$1.75}  & 8.08\tiny{$\pm$0.47}  & 10.03\tiny{$\pm$0.80}\\
    Proto           
    & 32.49\tiny{$\pm$2.01}   & 3.89\tiny{$\pm$0.24} & 10.68\tiny{$\pm$1.40} & 6.67\tiny{$\pm$0.46} & 13.43\tiny{$\pm$1.03} & 50.06\tiny{$\pm$1.57}
            &  9.54\tiny{$\pm$0.44} &  17.26\tiny{$\pm$2.65}  & 13.59\tiny{$\pm$1.61}  & 22.61\tiny{$\pm$1.57}\\
    De-MAML$^{\dag}$&42.40\tiny{$\pm$0.93} &4.76\tiny{$\pm$0.08} &23.87\tiny{$\pm$1.49}&14.19\tiny{$\pm$1.33}& 21.31\tiny{$\pm$0.96}& 37.10\tiny{$\pm$2.65} &6.21\tiny{$\pm$0.26} &18.54\tiny{$\pm$0.89}&26.41\tiny{$\pm$0.48}  &22.07\tiny{$\pm$1.07} \\
    L-TapNet+CDT&\textbf{44.30\tiny{$\pm$3.15}} &\textbf{12.04\tiny{$\pm$0.65}} &20.80\tiny{$\pm$1.06}&15.17\tiny{$\pm$1.25}& 23.08\tiny{$\pm$1.53} & 45.35\tiny{$\pm$2.67} &11.65\tiny{$\pm$2.34} &23.30\tiny{$\pm$2.80}& 20.95\tiny{$\pm$2.81} & 25.31\tiny{$\pm$2.65}\\
            Ours& 41.21\tiny{$\pm$0.70}& 9.91\tiny{$\pm$0.50} & \textbf{25.01\tiny{$\pm$1.32}}&\textbf{26.18\tiny{$\pm$2.69}} & \textbf{25.58\tiny{$\pm$1.30}} & \textbf{51.97\tiny{$\pm$2.93}} &\textbf{15.65\tiny{$\pm$1.51}} & \textbf{29.89\tiny{$\pm$1.64}} &\textbf{37.19\tiny{$\pm$1.35}} & \textbf{33.68\tiny{$\pm$1.86}}\\
    \bottomrule
    \end{tabular}}
\end{table*}

\subsection{Results (RQ1)}
We conduct an empirical study to investigate whether our proposed model can achieve better performance for few-shot sequence labeling to answer RQ1.
Table~\ref{fewnerd} and Table~\ref{cross-dataset} report the performance of our proposed model and baselines on few-shot NER dataset (i.e., FewNERD and Cross datasets) and Table~\ref{tab:snips} reports the performance on few-shot slot tagging dataset (i.e., SNIPS dataset).

First, we observe the performance on the FewNERD dataset.
We can see that our model consistently outperforms all baselines under all settings.
Compared with the best baseline ESD, our model achieves 3.68 and 2.58 average F1 improvement under Intra and Inter settings.
Compared with the best token-level baseline StructShot, our model achieves 14.85 and 13.18 average F1 improvement under two settings.
This shows that combining token-level and span-level can achieve better performance for few-shot sequence labeling.
It is worth noting that the improvement under Intra setting is higher than under Inter setting.
This suggests that when the domain gap between source domain and target domain is relatively large, combining token and span levels can lead to more significant improvements than the single-granularity baselines.

Secondly, we observe the performance on the SNIPS dataset.
Compared with the best baseline ESD, our model achieves 1.12 and 1.38 average F1 improvement under 1-shot and 5-shot settings on the SNIPS dataset.
Compared to all baseline methods, our method achieves the best or second best performance in most domains.
The results also show the effectiveness of our model.
It is worth noting that the standard deviations of our method are the best or second best in most settings.
This shows that our model is robust.

Thirdly, we observe the performance on the Cross dataset.
Compared with the best baseline L-TapNet+CDT, our model achieves 2.5 and 8.37 average F1 improvement under 1-shot and 5-shot settings.
Compared to all baseline methods, our method achieves the best or second best performance in all domains.
This shows the effectiveness of our model.

\begin{table}
\centering
\caption{The ablation study of our model on the development set on FewNERD dataset under 5-way 1\textasciitilde2 shot setting. The ablation study is repeated five times.}  \label{tab:ablation}
\begin{tabular}{lcccc}
\toprule
\textbf{Model Setting}&\textbf{Intra} &\textbf{Inter}\\
\midrule
\textbf{Ours}& \textbf{36.92\tiny{$\pm$1.54}} & \textbf{48.63\tiny{$\pm$0.51}} \\
\midrule
\textbf{w/o Span-Level Loss}  & 26.42\tiny{$\pm$0.33}  & 36.25\tiny{$\pm$1.21}\\
\textbf{w/o Token-Level Loss} & 34.53\tiny{$\pm$0.51}  & 47.80\tiny{$\pm$0.64}\\
\midrule
\textbf{w/o Consistent Loss} & 35.27\tiny{$\pm$0.77}   & 48.10\tiny{$\pm$0.16}  \\
\midrule
\textbf{w/o Inference Algorithm}& 31.13\tiny{$\pm$1.66} & 45.18\tiny{$\pm$0.31}  \\
\midrule
\textbf{w/o Division of O}& 35.63\tiny{$\pm$0.94} & 48.49\tiny{$\pm$0.40}  \\
\textbf{w/o Adaptive Prototype} & 32.52\tiny{$\pm$1.31}   & 46.27\tiny{$\pm$1.23}  \\
\textbf{w/o Cross-Attention}& 32.51\tiny{$\pm$1.78} & 44.98\tiny{$\pm$0.31}  \\
\bottomrule
\end{tabular}
\end{table}

\subsection{Ablation Study (RQ2)}
To validate the contribution of each component, we report the performance of removing each of them from our model individually in Table~\ref{tab:ablation} to answer RQ2.

First, we can see that all components improve performance.
This shows that all components in our model are effective.
Secondly, we remove the token-level loss and span-level loss.
When we ablate span loss, the network uses the output of the token-level network as the result.
We can see that the performance drops to 26.42 and 36.25 under Intra and Inter settings.
This may be because token-level metric learning is difficult to extract entities completely and our token-level model is parameter-efficient.
When we remove the token-level loss, the performance drops to 34.53 and 47.80 under two settings.
This shows that using token-level labels to jointly train the model can effectively improve performance.
Thirdly, we remove the consistent loss and the performance drops to 35.27 and 48.10 under two settings.
This shows that consistent loss can effectively regularize the model and improve performance.
Fourth, we remove the inference algorithm, i.e., all spans that are predicted as entities or slots from span-level network are extracted.
The performance drops to 31.13 and 45.18 under two settings.
This illustrates that our inference algorithm can effectively combine the results of two networks and avoid extracting overlapping spans.
Finally, we ablate all components (i.e., fine-grained division of O, cross-attention module, and adaptive prototype) in the span network.
We can see that all components in the span network improve performance.
The cross-attention module brings the largest performance improvement.
This indicates that the interaction of the support set and query set is the key to performance improvement.
We remove the adaptive prototype and the performance drops to 32.52 and 46.27 under two settings.
This shows adaptively constructing prototypes for each query span is more effective.
We remove the division of class O and the performance drops to 35.63 and 48.49 under two settings.
This shows that the fine-grained division of O can effectively construct prototype of class O.

\begin{table} 
\centering
\caption{The performance of our model under different inference algorithms on the development set of FewNERD dataset under Inter 5-way 1\textasciitilde2 shot setting. The experiment is repeated five times.}  \label{tab:inference}
\begin{tabular}{lcccc}
\toprule
\textbf{Inference Setting} &\textbf{F1}\\
\midrule
\textbf{Ours}  & \textbf{48.63\tiny{$\pm$0.51}}\\
\midrule
\textbf{Token Network} & 26.94\tiny{$\pm$0.96}\\
\textbf{Span Network}  & 48.05\tiny{$\pm$0.54}\\
\midrule
\textbf{Intersection} & 29.93\tiny{$\pm$0.71}\\
\textbf{Union}  & 41.27\tiny{$\pm$0.36}\\
\bottomrule
\end{tabular}
\end{table}

\subsection{Effects of Different Inference Algorithms (RQ3)}
We explore the effects of different inference algorithms in Table~\ref{tab:inference} to answer RQ3, including our proposed consistent greedy inference algorithm, using intersection or union operation to combine two outputs from two networks (i.e., a span and its type are extracted by both networks or by one), and using the outputs of token network or span network (i.e., we also use greedy algorithm to avoid overlapping spans).

First, we can see that our proposed inference algorithm achieves the best performance.
This shows that using token-level predictions to adjust the probability of span network is effective.
Secondly, we can see that the performance of span network is better than token network.
This may be because token-level metric learning is difficult to extract entities completely and our token-level model is parameter-efficient.
Thirdly, we report the performance of using intersection or union operation to combine the outputs of two networks.
We can see that their performance does not exceed span network.
This is because the poor performance of token network and simple combination does not improve performance.
Finally, it is worth noting the performance of span network exceeds w/o span-level loss.
This shows that using token-level loss to jointly train can improve performance of span-level network.

\subsection{Unidirectional vs. Bidirectional Consistent Loss (RQ4)}
We explore the performance of using unidirectional Kullback-Leibler divergence with temperature as consistent loss in Table~\ref{loss} to answer RQ4, i.e., only using Span network to supervise Token network (ST) or using Token network to supervise Span network (TS).

First, we can see that the bidirectional consistent loss performs better than ST and TS.
This shows that the bidirectional consistent loss is more effective to align outputs from two networks compared to the unidirectional consistent loss.
Secondly, ST and TS both exceed w/o consistent loss.
This shows that both unidirectional consistent losses are effective to align outputs and improve performance.
Thirdly, the performance of TS exceeds ST.
This suggests that it is more effective to use supervision information from token network to train the span network.
This may be because it is easier for span network to learn a consistent output.

\subsection{Effects of Different Consistent Losses (RQ4)}
We explore the performance of using different losses as consistent loss in Table~\ref{loss} to answer RQ4, i.e., mean squared error (MSE), Kullback-Leibler divergence (KL), negative cosine similarity (COS), and Jensen-Shannon divergence (JS).

First, we can see that the KL loss with temperature performs best and the KL loss performs second best.
This shows that using temperature can obtain better supervision information compared to KL loss.
It is worth noting that KL divergence and JS divergence exceed w/o consistent loss.
This shows that KL divergence and JS divergence can effectively measure the difference between two distributions and improves performance for our model.
Secondly, the performance of w/o consistent loss exceeds MSE and COS.
This illustrates that MSE and COS play a side effect and are not a good choice for measuring the difference between the two probability distributions for our model.

\begin{table} \setlength{\tabcolsep}{12pt}
\centering
\caption{The performance of our model under different loss settings on the development set of FewNERD dataset under Inter 5-way 1\textasciitilde2 shot setting. ST denotes using span network to supervise token network and TS denotes using token to supervise span. MSE, KL, COS, and JS denote mean squared error, Kullback-Leibler divergence, negative cosine similarity, and Jensen-Shannon divergence. The experiment is repeated five times.}  \label{loss}
\begin{tabular}{lcccc}
\toprule
\textbf{Consistent Loss} &\textbf{F1}\\
\midrule
\textbf{Ours}& \textbf{48.63\tiny{$\pm$0.51}}\\
\midrule
\textbf{ST} &  48.19\tiny{$\pm$0.49}\\
\textbf{TS} &  48.41\tiny{$\pm$0.44}\\
\midrule
\textbf{MSE}  & 48.05\tiny{$\pm$0.32}\\
\textbf{KL}   & 48.41\tiny{$\pm$0.13}\\
\textbf{COS}  & 47.93\tiny{$\pm$0.14}  \\
\textbf{JS}   & 48.38\tiny{$\pm$0.46}\\
\bottomrule
\end{tabular}
\end{table}

\begin{figure*}
  \centering
  \subfigure[Effects of $\lambda$]{
        \includegraphics [width=0.32 \textwidth]{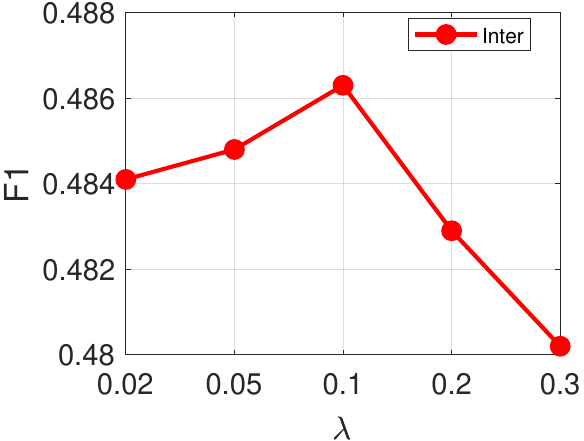}}
  \subfigure[Effects of $\lambda$]{
        \includegraphics [width=0.32 \textwidth]{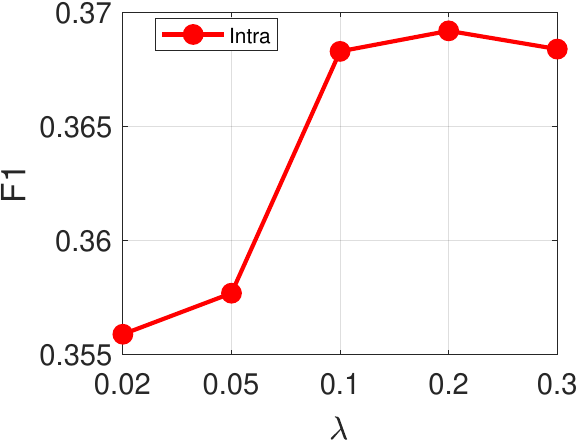}
  }
  \subfigure[Effects of $\gamma$]{
         \includegraphics [width=0.32 \textwidth]{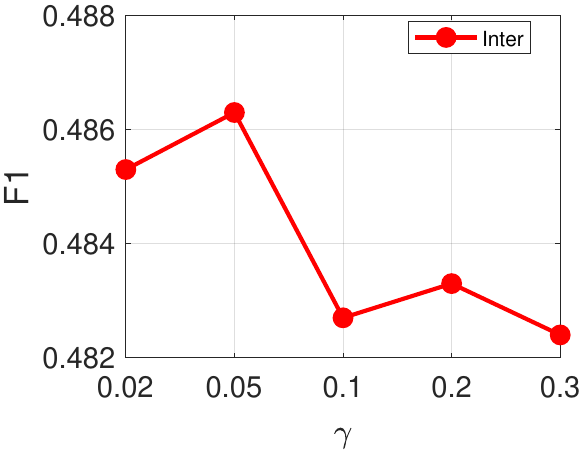}}
  \subfigure[Effects of $\gamma$]{
         \includegraphics [width=0.315 \textwidth]{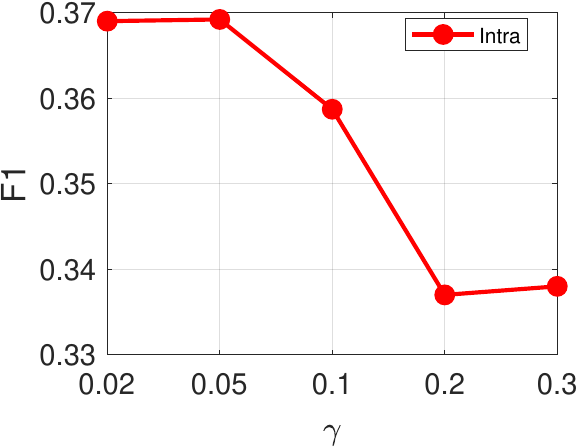}
  }
  \subfigure[Effects of $\delta$]{
         \includegraphics [width=0.32 \textwidth]{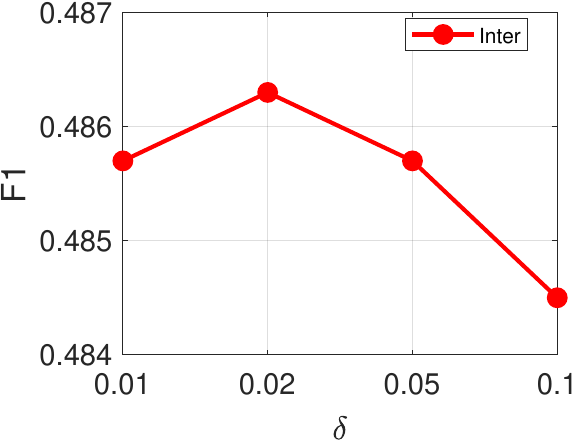}}
   \subfigure[Effects of $\delta$]{      
         \includegraphics [width=0.32 \textwidth]{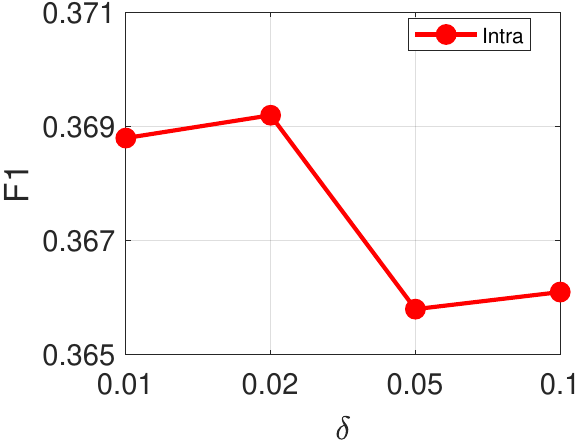}
  }
  \caption{Effects of Hyper-Parameters under 5-way 1\textasciitilde2 shot setting.} \label{hyper}
\end{figure*}

\subsection{Effects of Hyper-Parameters (RQ5)}
We explore the performance of different hyper-parameters (i.e., the weight of token-level loss $\lambda$ in Eq. (18), the weight of consistent loss $\gamma$ in Eq. (18), and the degree of penalty in inference algorithms $\delta$ in Eq. (19)) while fixing the other hyper-parameters in Fig.~\ref{hyper} on FewNERD dataset under 5 way 1\textasciitilde2 shot setting to answer RQ5.

We first observe the effects of $\lambda$ which is selected from \{0.02, 0.05, 0.1, 0.2, 0.3\}.
From Figure~\ref{hyper}(a) and Figure~\ref{hyper}(b), we find that the performance is relatively stable under Inter and Intra settings, i.e., the performance fluctuates by 0.6\% and 1.5\%, respectively.
As $\lambda$ changes, the performance first has an upward trend and then has a downward trend under two settings.
This shows that setting a suitable $\lambda$ can help improve performance.
The optimal value of $\lambda$ depends on the specific setting.
The model achieves the best performance when $\lambda$ is 0.1 under Inter setting and $\lambda$ is 0.2 under Intra setting.

Then, we observe the effects of $\gamma$ and $\delta$, choosing from \{0.02, 0.05, 0.1, 0.2, 0.3\} and \{0.01, 0.02, 0.05, 0.1\}.
The overall trend is the same as $\lambda$.
We can see that the performance is stable under two settings, i.e., the performance fluctuates by 0.5\% and 0.4\% with the change of $\gamma$ and the performance fluctuates by 0.2\% and 0.3\% with the change of $\delta$.
This shows that our model is insensitive to $\gamma$ and $\delta$.
As $\gamma$ or $\delta$ changes, the overall performance also first has an upward trend and then has a downward trend under two settings.
The model achieves the best performance when $\gamma$ is 0.05 and $\delta$ is 0.02 under two settings.

\subsection{Comparison with Other O Divisions}

\begin{table}
\centering
\caption{The performance of different O divisions on the development set of FewNERD dataset under Inter 5-way 1\textasciitilde2 shot. The upper part of the table shows the performance of our model under different divisions, and the lower part shows the ESD model under different divisions. All results over five runs are reported.} \label{partition}
\begin{tabular}{lccccc}
\toprule
\textbf{Model Setting} & \textbf{Division Setting} &\textbf{F1}\\
\midrule
\multirow{2}{1cm}{\tabincell{c}{\textbf{CDAP}}}&
\textbf{Ours} &  \textbf{48.63\tiny{$\pm$0.51}} \\
&\textbf{\citet{DBLP:conf/naacl/WangXLZCCS22}}  & 48.39\tiny{$\pm$0.68} \\
\midrule
\multirow{2}{1cm}{\tabincell{c}{\textbf{ESD}}}&
\textbf{Ours}  &\textbf{46.55\tiny{$\pm$0.41}} \\
&\textbf{\citet{DBLP:conf/naacl/WangXLZCCS22}} & 45.83\tiny{$\pm$0.56}\\
\bottomrule
\end{tabular}
\end{table}

We compare our division with~\citet{DBLP:conf/naacl/WangXLZCCS22} under two different models (i.e., CDAP and ESD) in Table~\ref{partition} to show the effectiveness of our division.
\citet{DBLP:conf/naacl/WangXLZCCS22} also divide class O into three sub-classes: sub-class $O_1$ denotes the span that does not overlap with any entities or slots in the sentence, sub-class $O_2$ denotes the span that is the sub-span of an entity or slot, and $O_3$ denotes the others.

We can see that our O division achieves better performance under two different models.
This illustrates the effectiveness of dividing sub-classes according to the left and right boundary tokens.
It is worth noting that our division is more effective on the ESD model.
This may be because the capacity of distinguishing sub-classes for the ESD model is limited compared to our model, and our division can help ESD model to distinguish sub-classes more effectively.

\subsection{Efficiency}
Since our model considers both token-level and span-level, we compare our method with token-level methods (i.e., Proto, NNShot, and StructShot) and span-level methods (i.e., ESD and De-MAML) in terms of the number of parameters and inference time in Table~\ref{tab:eff} to show the efficiency.

First, we observe the number of parameters and we can see that Proto, NNShot, and StructShot have the least number of parameters.
This is because these methods only use BERT to encode the tokens.
Our model is similar to ESD and significantly smaller than De-MAML.
This shows that our method is parameter-efficient and does not introduce too many parameters (i.e., 2M) compared to Proto.
In addition, De-MAML uses two BERT to encode two tasks separately resulting in a huge number of parameters.

Secondly, we compare the inference time of our model and baselines with the same hardware and batch size (i.e., 1).
We can see that the token-level methods are faster compared to the span-level methods.
This is because the span-level methods need to enumerate all spans within a sentence whose length is less than or equal to L.
Therefore, the number of spans is approximately L times that of tokens, which may bring extra computation overhead.
Proto is faster compared to NNShot, and NNShot is faster compared to StructShot.
This is because Proto only needs to measure the distance to the prototype, while NNShot needs to measure the distance to all the tokens in the support set.
StructShot further uses a Viterbi decoder to model label dependency based on NNShot, so more inference time is required.
Then, we compare our model with span-level methods and we can see that our model is 13.40 ms slower than ESD and 32.49 ms faster than De-MAML under 1-shot setting.
Under 5-shot setting, our model is 26.78 ms slower than ESD and 23.44 ms slower than De-MAML.
This illustrates that the time cost of our model is acceptable and our model is efficient.
In addition, De-MAML needs to forward propagate 2 times through BERT, resulting in a long inference time under 1-shot setting.
Our model and ESD both use the cross-attention module to model the interaction of the support set and query set, resulting in a significant increase in inference time under 5-shot setting.

\begin{table}[t]\setlength{\tabcolsep}{10pt}
\centering
\caption{The number of parameters and inference time per episode task on FewNERD dataset under Inter 5-way setting. F1 notes the micro-F1 over eight settings on the FewNERD dataset. \# Para. denotes the number of parameters and $\dag$ denotes that we do not fine-tune model during meta-testing phase for time efficiency.}
\begin{tabular}{lcccc}
\toprule
\multirow{2}{*}{\textbf{Models}}& \multirow{2}{*}{\textbf{F1}} & \multirow{2}{*}{\textbf{\# Para.}} & \multicolumn{2}{c}{\textbf{Inference Time}} \\ 
\cmidrule(r){4-5} 
~    & ~   &   & \textbf{1\textasciitilde2 shot}    & \textbf{5\textasciitilde10 shot}   \\    
\midrule
\textbf{Ours}  & 53.74  &111M      & 79.43 ms   & 234.23 ms         \\
\midrule
\textbf{Proto}    & 37.10 &  109M      & 24.98 ms   & 29.83 ms         \\
\textbf{NNShot}   & 37.37 & 109M     & 25.43 ms   & 30.97 ms         \\
\textbf{StructShot} & 39.73  & 109M  & 30.70 ms   & 49.62 ms         \\
\textbf{ESD}   & 50.61 &111M      & 66.03 ms   & 207.45 ms         \\
\textbf{De-MAML}$^{\dag}$ & 44.95 &  218M    & 111.92 ms   & 210.79 ms         \\ 
\bottomrule
\end{tabular}
\label{tab:eff}
\end{table}

\begin{table}
\centering
\caption{Error analysis on the test set of FewNERD dataset under 5-way 1\textasciitilde2 shot setting. \# FP-Span denotes the number and proportion of extracted entities with wrong span boundary. \# FP-Type denotes the number and proportion of extracted entities with the right boundary but the wrong entity type.}  \label{error}
\begin{tabular}{lcccc}
\toprule
\textbf{Setting} &\textbf{F1}&\textbf{\# FP-Span}&\textbf{\# FP-Type}\\
\midrule
\textbf{Inter}& 62.79 & 7372(74.52\%)  & 2619(26.48\%)\\
\textbf{Intra}& 41.21 & 11264(65.46\%) & 5944(34.54\%)\\
\bottomrule
\end{tabular}
\end{table}

\subsection{Error Analysis}
We further analyze the predicted error of our model in detail and divide the error into 2 categories (i.e., FP-Span and FP-Type) according to whether the span is correct.
FP-Span denotes extracted entities with the wrong span boundary and FP-Type denotes extracted entities with the right boundary but the wrong entity type.

First, as shown in Table~\ref{error}, we can see that FP-Span is main prediction error under two settings.
FP-span accounts for 74.52\% under Inter setting and 65.46\% under Intra setting.
This indicates that span detection is more difficult than span classification, which is the main bottleneck at present and needs further attention in the future.
Secondly, compared to Inter setting, the number of both errors rises under the more difficult Intra setting.
Compared to Inter setting, FP-Span increased by 52.79\% and FP-Type increased by 126.96\% under Intra setting.
This suggests that as domain difference between the source domain and the target domain increases, entity classification is more difficult to transfer than span detection.

\subsection{Case Study}
Apart from the quantitative analysis, we also conduct case study to intuitively show the effectiveness and analyze our model.
We list the predictions of two groups on the test set of our model and three strong baselines (i.e., StructShot, De-MAML, and ESD) on FewNERD under Inter 5-way 1\textasciitilde2 shot setting in Table~\ref{case}.

First, we observe the predictions of the first group (Query 1, 2, and 3).
We can see that the extraction capacity of StructShot is limited.
For example, it only correctly extracts entity ``B.A.'' in Query (1) and also extracts impossible entities (i.e., ``the'' and ``of'' in Query (3)).
This illustrates that StructShot relies heavily on the labeled samples in the support set.
Then, we observe the predictions of span-level methods.
For query (1), De-MAML and ESD predict None, and our model extracts ``B.A.''.
This may be because we use token-level loss to jointly train the model.
For query (2), we can see that our model correctly extracts and predicts ``Monaco'' and ``Maxim's''.
De-MAML fails to classify ``Maxim's'' and ESD fails to extract ``Maxim's''.
This shows that even if a sentence has multiple entities, our model can extract them.
For query (3), we can see that our model correctly extracts and predicts ``faculty of medicine in Paris''.
De-MAML and ESD extract ``faculty of medicine'', and ESD predicts wrong type.
This shows that even if an entity has multiple tokens, our model can locate and extract it.

Secondly, we display some failure examples in the second group (Query 4, 5, and 6).
For query (4), method StructShot extracts ``Right'' and fails to extract simple entity ``Republican''.
This also shows that the extraction capacity of StructShot is limited and relies heavily on the labeled samples in the support set.
Methods De-MAML, ESD, and our model do not extract entity ``neo-Confederate Right'' at all.
This may be because it is more difficult to extract this entity.
For query (5), we can see that StructShot does not extract the entity completely, only ``bachelor'' is extracted.
De-MAML and our model also do not extract the entity completely, only ``bachelor of law'' is extracted.
This indicates that there are challenges in locating the boundaries of entities.
ESD does not extract entities, perhaps because it is trained with only span labels and the supervision is sparse.
For query (6), we can see that all the methods predict None except De-MAML and De-MAML predicts wrong entity ``New American''.
We observe the corresponding support set and find that it may be due to the weak representation of the prototype constructed from the two samples (i.e., Bobino and Gaumont-Palace) in the support set of the class.

\begin{table}[t] \small
\definecolor{cGreen}{RGB}{41,163,95}
\definecolor{capri}{rgb}{0.0, 0.75, 1.0}
\caption{The predictions on test set of our proposed model and model variant for case study on FewNERD dataset under Inter 5-way 1\textasciitilde2 shot. We use different colors to indicate the different entity types in a sentence. \textcolor{capri}{blue} indicates that this entity type does not appear in this query sample.} \label{case}
\begin{tabular}{p{6.5cm}|p{1.6cm}|p{1.5cm}|p{1.3cm}|p{1.2cm}}
\toprule
\textbf{Query} & \textbf{StructShot}&\textbf{De-MAML}&\textbf{ESD}&\textbf{Ours}\\
\midrule
(1) He decided to return to school in 1996, earning his {\color{cGreen}B.A. [other-educational degree]}.& {\textcolor{cGreen}{earning}}, {\textcolor{cGreen}{B.A.}} & None&None & {\textcolor{cGreen}{B.A.}}\\
\midrule
(2) The same year, {\color{orange}Monaco [person-actor]} was named ``{\color{cGreen}Maxim's [art-written art]}'' number one sexiest cover model of the decade.  & None& {\textcolor{orange}{Monaco}}, {\textcolor{capri}{Maxim's}} &{\textcolor{orange}{Monaco}} & {\textcolor{orange}{Monaco}}, {\textcolor{cGreen}{Maxim's}}  \\
\midrule
(3) In 1864 he attained the chair of internal pathology at the {\color{cGreen}faculty of medicine in Paris [building-hospital]}. &{\textcolor{cGreen}{the}}, {\textcolor{capri}{chair}}, {\textcolor{cGreen}{of}}, {\textcolor{cGreen}{of medicine in}}& {\textcolor{cGreen}{faculty of medicine}}& {\textcolor{capri}{faculty of medicine}} & {\textcolor{cGreen}{faculty of medicine in Paris}}\\
\midrule
\midrule
(4) She writes that ``even into the twenty-first century mainstream conservative {\color{blue}Republican [organization-political party]} politicians continued to associate themselves with issues, symbols, and organizations inspired by the {\color{orange}neo-Confederate Right [organization-political party]}''. &{\textcolor{orange}{Right}} & {\textcolor{orange}{Republican}}& {\textcolor{orange}{Republican}} & {\textcolor{orange}{Republican}}\\
\midrule
(5) He gained a {\color{cGreen}bachelor of law degree [other-educational degree]}. & {\textcolor{cGreen}{bachelor}}& {\textcolor{cGreen}{bachelor of law}} & None &{\textcolor{cGreen}{bachelor of law}}\\
\midrule
(6) The script was published in {\textcolor{orange}{Grove New American Theater [building-theater]}}. & None &{\textcolor{orange}{New American}} &None &None\\
\bottomrule
\end{tabular}
\end{table}

\section{Conclusion and Future Work} \label{sec:conclu}
In this paper, we first unify token and span level supervisions and propose a consistent dual adaptive prototypical network for few-shot sequence labeling.
Our model contains token-level network and span-level network, jointly trained with token-level and span-level labels.
Further, we propose a consistent loss to make the outputs of two networks consistent based on bidirectional Kullback-Leibler divergence with temperature.
During the inference phase, we propose a consistent greedy inference algorithm that first adjusts the predicted probability of span network and then greedily selects non-overlapping spans with maximum probability based on adjusted probability.
Extensive experiments show that unifying token and span level supervisions is effective and our model achieves new state-of-the-art results on three benchmark datasets.

In the future, we plan to deepen and widen our work from the following aspects:
(1) In this work, we first unify token and span level supervisions for few-shot sequence labeling and conduct a preliminary exploration.
The proposed method is demonstrated to be simple and effective, but whether other complicated token-level and span-level networks are also effective has not been explored, which is worthy of further exploration.
(2) The proposed method has only been verified to be effective on texts in English. 
The effectiveness on datasets of other languages deserves further verification.
(3) The effectiveness of our model on traditional supervised sequence labeling also deserves further exploration.
\bibliographystyle{ACM-Reference-Format}
\bibliography{sample-base}

\end{document}